\documentclass[10pt,twocolumn,letterpaper]{article}

\usepackage[pagenumbers]{wacv} % To force page numbers, e.g. for an arXiv version

\usepackage{graphicx}
\usepackage{amsmath}
\usepackage{amssymb}
\usepackage{booktabs}
\usepackage[normalem]{ulem}

\usepackage[pagebackref,breaklinks,colorlinks]{hyperref}

\usepackage[capitalize]{cleveref}
\crefname{section}{Sec.}{Secs.}
\Crefname{section}{Section}{Sections}
\Crefname{table}{Table}{Tables}
\crefname{table}{Tab.}{Tabs.}

\usepackage{graphicx}
\usepackage{amsmath}
\usepackage{amssymb}
\usepackage{mathrsfs}%\mathsf{}
\usepackage{mathtools}%\coloneq
\usepackage{booktabs}
\usepackage{rotating}
\usepackage{multirow}% multirow table
\usepackage{booktabs}% tables
\usepackage{lipsum}
\usepackage[dvipsnames,table]{xcolor}
\usepackage{enumitem}
\usepackage{inconsolata}
\usepackage{pifont}
\usepackage{soul}
\usepackage{siunitx}
\usepackage[export]{adjustbox}
\usepackage{balance}
\usepackage[accsupp]{axessibility}  % Improves PDF readability for those with disabilities.

\renewcommand{\paragraph}[1]{\vspace{1mm}\noindent\textbf{#1}}
\newcommand{\bbf}{\mathbf{f}}

\newcommand{\bg}{\mathbf{g}}
\newcommand{\bh}{\mathbf{h}}

\newcommand{\bW}{\mathbf{W}}
\newcommand{\bE}{\mathbf{E}}
\newcommand{\bbF}{\mathbf{F}}

\newcommand{\CLS}{\mathsf{CLS}}
\newcommand{\bbR}{\mathbb{R}}

\newcommand{\ML}{\mathcal{L}}

\newcommand{\tablevenue}[1]{\scriptsize{\color{gray} #1}}
\newcommand{\meansem}[2]{#1 {\scriptsize${\pm}$#2}}

\def\wacvPaperID{434} % *** Enter the WACV Paper ID here
\def\confName{WACV}
\def\confYear{2025}

\begin{document}

%%%%%%%%% TITLE - PLEASE UPDATE
\title{Seeing Eye to AI: \\ Comparing Human Gaze and Model Attention in Video Memorability}

\author{
Prajneya Kumar$^{*1}$ \hspace{2mm}
Eshika Khandelwal$^{*2}$ \hspace{2mm}
Makarand Tapaswi$^{\dagger 2}$ \hspace{2mm}
Vishnu Sreekumar$^{\dagger 1}$ \\
\{$^1$Cognitive Science Lab, $^2$CVIT\}, IIIT Hyderabad \\
{\small $^{* \dagger}$ equal contribution}
% For a paper whose authors are all at the same institution,
% omit the following lines up until the closing ``}''.
% Additional authors and addresses can be added with ``\and'',
% just like the second author.
% To save space, use either the email address or home page, not both
% \and
% Second Author\\
% Institution2\\
% First line of institution2 address\\
% {\tt\small secondauthor@i2.org}
}

\twocolumn[{
\renewcommand\twocolumn[1][]{#1}%
\maketitle
\centering
\vspace{-5mm}
\includegraphics[width=\linewidth]{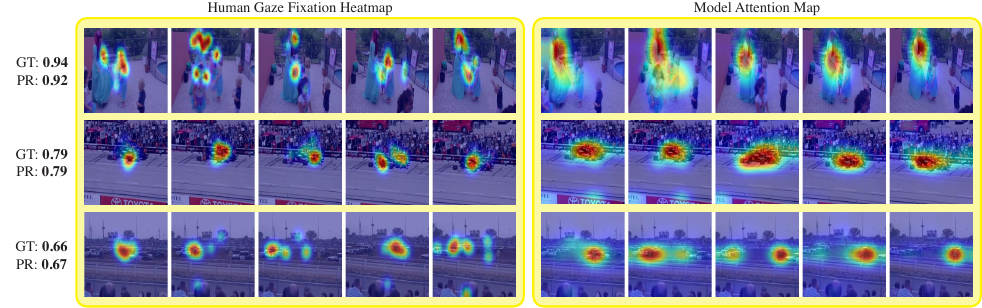}
\vspace{-6mm}
\captionof{figure}{Comparing human gaze fixations (left) and model's attention maps (right) for 3 different videos (one per row). 
The memorability scores, ground-truth (GT) and  model prediction (PR), are provided on the left.
The heatmaps depict areas of high visual attention through warmer colors (red-yellow), indicating regions where human observers fixated (left) and model attended (right).
The model's attention patterns are aligned with human gaze patterns, especially for more memorable videos.
Samples from Memento10k~\cite{Newman2020}.
}
\label{fig:teaser}
\vspace{6mm}
}]

\vspace{-2mm}
\begin{abstract}
\vspace{-3mm}
Understanding what makes a video memorable has important applications in advertising or education technology.
Towards this goal, we investigate spatio-temporal attention mechanisms underlying video memorability.
Different from previous works that fuse multiple features, we adopt a simple CNN+Transformer architecture that \emph{enables analysis of spatio-temporal attention} while matching state-of-the-art (SoTA) performance on video memorability prediction.
We compare model attention against human gaze fixations collected through a small-scale eye-tracking study where humans perform the video memory task.
We uncover the following insights:
(i)~Quantitative saliency metrics show that our model, trained only to predict a memorability score, exhibits similar spatial attention patterns to human gaze, especially for more memorable videos.
(ii)~The model assigns greater importance to initial frames in a video, mimicking human attention patterns.
(iii)~Panoptic segmentation reveals that both (model and humans) assign a greater share of attention to \textit{things} and less attention to \textit{stuff} as compared to their occurrence probability.

\end{abstract}

\vspace{-2mm}
\section{Introduction}
\label{sec:intro}

In 2018, Nike's ``Dream Crazy'' commercial featuring Colin Kaepernick captured nationwide attention in the US\footnote{Dream Crazy \url{https://www.youtube.com/watch?v=WW2yKSt2C_A}}.
This advertisement was especially memorable because it was aired in the aftermath of Kaepernick's protests against race-based police brutality. While the context made this commercial memorable for US-based audiences, other types of commercials tend to be memorable in general.
For example, a famous 2013 E-Trade Super Bowl commercial features a baby seated behind a stack of cash talking about investments and hidden fees\footnote{E-Trade ad \url{https://www.youtube.com/watch?v=EbnWbdR9wSY}}.
This sort of ad is likely to be memorable regardless of cultural context due to several attention-grabbing features, notably, a baby talking in an adult voice and delivering investment advice.
This latter type of memorability, thought to be consistent across individuals and cultures, has been extensively studied in both cognitive science and computer vision
using images~\cite{bainbridge_chapter_2019, Isola2011} and words~\cite{aka_semantic_2023, Madan2021}. 
In this work, we ask: what are the spatial, temporal, and semantic patterns of attention that are associated with video memorability?
To answer this question, we train a CNN+Transformer model to predict human memorability of naturalistic videos,
use self-attention scores to determine where the model \emph{looks} across space and time,
and collect human eye-tracking data to compare the model's attention against human fixations (\cref{fig:teaser}).

Early work on image memorability reveals the importance of both object and scene categories in predicting memorability~\cite{Isola2011,Dubey2015}.
Semantic categories are also predictive of memorability across stimuli, including words~\cite{aka_semantic_2023,Madan2021} and indeed, prior work shows that context guides eye movements to task-relevant object locations~\cite{Torralba2006eyemov}.
Thus, we investigate what semantic categories in videos drive memorability.
Video captioning approaches have been used in previous semantic analyses of video memorability~\cite{Cohendet2018VideoCap, Newman2020, Shekhar2017ShowRecall}.
However, to our knowledge, we are the first to present a detailed analysis of attention captured by different semantic categories when humans attempt to memorize videos and when a model is trained to predict these memorability scores.
We apply panoptic segmentation~\cite{maskformer} and adopt the COCO hierarchy~\cite{Caesar2018coco} to distinguish between \emph{things} (\ie~objects with well-defined shapes such as \emph{person}) and \emph{stuff} (\ie~amorphous background regions such as \emph{sky}) in the video frames.
Next, we compare pixel distributions weighted by model attention and human gaze and find that both the model and humans generally enhance attention to \emph{things} and reduce attention to \emph{stuff}.
Furthermore, the model and humans agree on what specific \emph{things} and \emph{stuff} to emphasize or disregard.
Overall, these results indicate that the model learns similar attentional strategies as humans \textit{even though it is trained only to predict a memorability score}.

Beyond semantics, the time axis in videos begs an important question: how early does the model know about the memorability of a video?
Human experiments using extremely fast presentation times reveal that image memorability differences can be observed in brain activity patterns as early as \SI{400}{\milli\second}~\cite{bainbridge_chapter_2019,khaligh-razavi_what_2016}.
Therefore, it is possible that very early moments in a video are predictive of how memorable it will be.
Furthermore, human attention tends to be highest at the beginning of an event and wanes over the course of the event~\cite{kosie_attentional_2019}.
Thus, video memorability scores may be influenced to a greater extent by the initial frames.
Note that memorability scores are computed as a consensus across participants.
Therefore, we expect the video frames that most people attend to in similar ways to drive the memorability scores.
Despite having no intrinsic temporal bias, can models trained to predict memorability pick up on these human-like temporal attention patterns? 
To answer this question, we first analyze human-human gaze agreement in our videos and establish that different people are more likely to attend to similar regions in the initial frames.
Next, summing over the model's spatial attention scores in a frame, we observe that the model indeed assigns greater importance to earlier frames within videos, thereby discovering a subtle temporal pattern in human behavior.

The video memorability literature~\cite{Cohendet2018,Dumont2023,Harini2023} focuses on high prediction performance and lacks analysis of models' (dis)similarities to how humans view and remember videos.
We address this gap through the following contributions:
(i)~We adopt a simple CNN+Transformer model to predict video memorability as it facilitates a study of spatio-temporal attention mechanisms.
Even with a single encoder, our model matches state-of-the-art performance.
(ii)~To compare the model against \emph{what} humans look at and \emph{when}, we collect eye-tracking data of subjects in a video memorability experiment, similar to the original setup~\cite{Cohendet2018,Newman2020}.
(iii)~Through panoptic segmentation and attention-weighted analyses, we show that both the model and humans increase and decrease attention similarly to different \emph{things} and \emph{stuff}.
(iv)~We show that our model with no intrinsic temporal bias learns to attend to the initial frames of the video with a decreasing pattern over time, consistent with framewise human-human gaze agreement patterns.
We will release our code and eye-tracking data to encourage further research.

Note, our work aims to highlight the similarities between \textit{human fixations} when performing memorability experiments, and \textit{model attention} when trained to predict memorability scores.
A simple CNN+Transformer architecture enables this, matches SoTA, and has not been used in video memorability before.

\section{Related Work}
\label{sec:relwork}

\paragraph{Memorability in cognitive science.}
While human beings remember a huge amount of visual information, not all visual experiences are equal in our memory~\cite{Isola2011}.
Some images are consistently better remembered across people, suggesting that memorability is observer-independent~\cite{bainbridge_chapter_2019,bainbridge_memorability_2017}.
This makes algorithms suitable for predicting memorability~\cite{khosla_understanding_2015}.
Several factors such as scene semantics~\cite{Isola2011}, object category~\cite{Dubey2015}, and visual saliency~\cite{Dubey2015} correlate with memorability, yet considerable statistical variance in memorability scores remains unexplained~\cite{rust_understanding_2020}. 
Although image memorability has been studied extensively in cognitive science, videos have been used primarily in the study of event segmentation and to understand the neural processes underlying learning and memory~\cite{bird_consolidation_2015,baldassano_discovering_2017}.
Observer-independent memorability of videos has received less attention in cognitive science compared to the work in computer vision.

\paragraph{Memorability in computer vision.}
The study of visual memorability in computer vision started with a focus on images~\cite{Isola2011,khosla_understanding_2015}.
Models such as \emph{MemNet} were developed for image memorability prediction on large image datasets~\cite{khosla_understanding_2015}.
Improvements over the initial models involved incorporating
attention mechanisms~\cite{fajtl_amnet_2018},
image captioning modules~\cite{squalli2018Lamem},
object and scene semantics~\cite{perera_cvprw_2019}, and
aesthetic attributes~\cite{zhu_aesthetics-assisted_2020}.
The insights gained from these studies also led to the development of Generative Adversarial Networks (GAN) based models that can modify images to manipulate their memorability~\cite{Goetschalckx_GANalyze_2019,sidorov2019changing,kyle2020generating}. 

Video memorability has fewer works, typically evaluated on \emph{VideoMem}~\cite{Cohendet2018} and
\emph{Memento10k}~\cite{Newman2020}.
The semantic embeddings model of \emph{VideoMem}~\cite{Cohendet2018} uses an image-captioning pipeline in conjunction with a 2-layer MLP for memorability prediction.
\emph{SemanticMemNet}~\cite{Newman2020} integrates visual cues with semantic information and decay patterns to predict memorability.
Recent approaches involve multiple tiered representation structures, \emph{M3S}~\cite{Dumont2023}, or use Large Language Models (LLMs) to generate textual descriptions that are then used to predict memorability scores~\cite{Harini2023}.
In contrast, we adopt a simple CNN+Transformer attention-based model that matches SoTA, but also facilitates comparison between model attention and human gaze on semantic and temporal aspects of video memorability.

\section{Methods: Model and Human}
\label{sec:method}

\begin{figure}[t]
\centering
\includegraphics[width=0.95\linewidth]{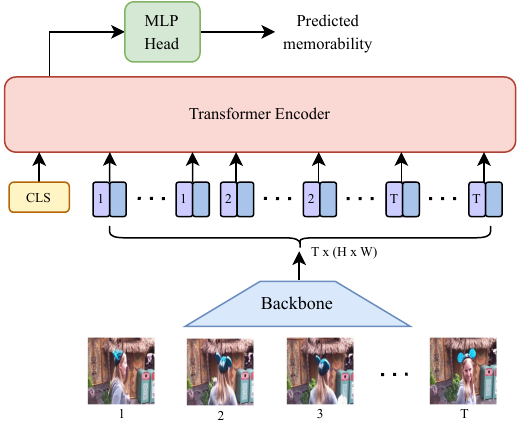}
\vspace{-1mm}
\caption{\textbf{Model overview.}
$T$ video frames are passed through an image backbone encoder to obtain spatio-temporal features $\bbF \in \bbR^{T \times H \times W \times D}$.
Coupled with position embeddings, and after appending a $\CLS$ token, we pass them through a Transformer encoder with self-attention.
A memorability score is calculated at the $\CLS$ representation with an MLP.
\textit{Attention scores between $\CLS$ and each token are used for downstream anaylsis}.}
\label{fig:model_arch}
\vspace{-3mm}
\end{figure}

We present two methods:
(i)~a CNN+Transformer model that predicts memorability scores using spatio-temporal attention; and
(ii)~an eye-tracking study to capture human gaze patterns during a memorability experiment.

\subsection{Transformer-based Model}
\label{subsec:cnn_transformer}

We begin by defining some notation.
Our dataset consists of multiple videos with associated memorability scores, $(V, m)$ pairs.
Each video consists of multiple frames.
We sub-sample $T$ frames for memorability prediction and denote a video as $V = \{f_i\}_{i=1}^{T}$.

Our model consists of three parts:
(i)~a backbone image encoder $\Phi$,
(ii)~a Transformer encoder that attends over spatio-temporal tokens extracted from $T$ video frames, and
(iii)~a prediction head that estimates the memorability of a video (see \cref{fig:model_arch}).

\paragraph{1. Image encoder.}
Our goal is to employ a model that allows us to analyze the spatio-temporal attention over video frames.
Thus, we consider CNN backbones such as ResNet-50~\cite{he2016resnet}, trained with contrastive language-image pretraining (CLIP)~\cite{radford2021clip}.
We encode each video frame to obtain a space-aware representation (from the \texttt{conv5} layer):
\begin{equation} 
\bbf_i = \Phi(f_i) \, , \text{ where } \bbf_i \in \bbR^{H \times W \times D} \, , \forall i \in \{1, \ldots, T\} \, ,
\end{equation}
where $H \times W$ are height and width of the spatial resolution, and $D$ is the dimensionality of the embeddings.

While previous works use multiple features:
frames, flow, and video by~\cite{Newman2020};
low-, mid-, and high-level representations and a contextual similarity module by~\cite{Dumont2023}; or
a host of 10+ models fed to an LLM by~\cite{Harini2023},
our model relies on a single semantic backbone (CLIP).
Our simple approach enables the analysis of model's spatio-temporal attention maps through a comparison to human gaze.

\paragraph{2. Video encoder.}
We use a Transformer encoder~\cite{vaswani2017transformer} to capture attention across spatio-temporal tokens.
First, we flatten and encode the image features using a linear layer $\bW_d \in \bbR^{d \times D}$ to reduce dimensionality.
Next, to each token, we add two types of position embeddings:
\begin{equation}
\small 
\bbf'_{ij} = \bW_d \bbf_{ij} + \bE^t_i + \bE^s_j \, , 
\forall i \in \{1, \ldots, T\}, 
j \in \{1, \ldots, HW\} \, ,
\label{eq:pos_embed}
\end{equation}
where $\bE^t_i$ is the $i^\text{th}$ row of the temporal embedding matrix (learnable or Fourier), and
$\bE^s_j$ is the $j^\text{th}$ row of the spatial embedding matrix, and
$\bbf_{ij} \in \bbR^{D}$ is the feature at frame $i$ and spatial region $j$.

We prepend a $\CLS$ token (with learnable parameters $\bh_\CLS$) to create a sequence of $1{+}THW$ tokens and post LayerNorm~\cite{layernorm} 
feed this to a Transformer encoder (TE) of $L$ layers with hidden dimension $d$:
\begin{equation}
\small
[\tilde{\bh}_\CLS, \tilde{\bbf}_{11}, \ldots, \tilde{\bbf}_{THW}] = 
\text{TE}( [\bh_\CLS, \bbf'_{11}, \ldots, \bbf'_{THW}] ) \, .
\label{eq:te}
\end{equation}

\paragraph{3. Predicting memorability.}
We pass the $\CLS$ token's contextualized representation to an MLP and predict the memorability score:

$\hat{m} = \text{MLP}( \tilde{\bh}_\CLS )$.

\paragraph{Extracting attention scores.}
We extract the self-attention matrix from the multi-head attention module of the last layer of the TE.
We mean pool over the heads and pick the row corresponding to the $\CLS$ token.
Ignoring the self token, this attention vector $\alpha \in \bbR^{THW}, \sum \alpha = 1$, is used for further spatio-temporal analysis.
We obtain an attention map of the size of the image by applying upscaling (pyramid expand) on the $H\times W$ attention scores of each frame.

\paragraph{Training and inference.}
Similar to previous work~\cite{Newman2020, Dumont2023} we use the MSE loss $\ML = \| m - \hat{m} \|^2$ to train our model.
We also considered the Spearman loss~\cite{Dumont2023}, but did not see significant performance gains.
For most experiments, we freeze the backbone and rely on the strong semantic features extracted by CLIP pretraining.

\subsection{Eyetracking Study: Capturing Gaze Patterns}
\label{sec:eyetracking_study}

We collect eye-tracking data while participants view videos in a memory experiment.
The setup (schematic in supplement \cref{fig:continuous_recognition}) follows the original video memorability experiments~\cite{Cohendet2018, Newman2020}, as we want the gaze patterns to accurately reflect the cognitive and visual processes involved in viewing and remembering videos. Further details regarding the setup are provided in supplement \cref{sec:eyetracking_setup}.

\paragraph{Data collection.}
Our study has $20$ participants ($9$ females, $11$ males, Age \meansem{22.15}{0.52} (\meansem{mean}{SEM})).
Memento10K: $6$ females, $4$ males, Age \meansem{22.9}{0.94}.
VideoMem: $3$ females, $7$ males, Age \meansem{21.4}{0.37}.
We choose $140$ unique videos each from both video datasets:
\emph{Memento10K}~\cite{Newman2020} and \emph{Videomem}~\cite{Cohendet2018}.
We use the SR Research EyeLink 1000 Plus~\cite{SRResearch2023EyeLink} to capture binocular gaze data, sampling pupil position at \SI{500}{\hertz}.
A 9-point target grid is used to calibrate the position of the eye.
Saccades and fixations are defined using the algorithm supplied by SR Research.

We perform clustering to select videos spanning diverse visual content and memorability attributes (see supplement ~\cref{sec: eyetracking_procedure} for details).
Participants watch multiple videos and are instructed to press the \texttt{SPACEBAR} upon identifying a repeated video.
Each participant watched a total of $200$ videos:
$140$ unique videos,
$20$ target repeats occurring at an interval between $9{-}200$, and
$40$ vigilance repeats interspersed every $2-3$ videos.
All videos are displayed in their original aspect ratios at the center of a white display screen with resolution $1024 \times 768$ pixels.

\paragraph{Data processing.}
The fixation coordinates for both eyes are obtained using the EyeLink Data Viewer software package (SR Research Ltd., version 4.3.210). These coordinates are then used to construct a binary matrix for each participant, corresponding in size to the original video dimensions.
To account for the visual angle of approximately 1 degree, a Gaussian blur is applied to these matrices (see supplement ~\cref{sec:gaussian_blur} for details).
To create the human fixation density maps, we average the  matrices corresponding to the same frame of the same video across participants.
To ensure compatibility with model's attention maps, the fixation maps are resized to a resolution of $224 \times 224$ pixels.

\section{Experiments}
\label{sec:experiments}

\paragraph{Video memorability datasets.}
We perform experiments on two datasets:
(i)~\textbf{VideoMem}~\cite{Cohendet2018} consists of 10K, 7 second video clips, each associated with a memorability score.
(ii)~\textbf{Memento10K}~\cite{Newman2020}, introduced as a dynamic video memorability dataset, contains human annotations at different viewing delays.
This dataset consists of 10K clips, but they are shorter in duration (3 seconds).

\paragraph{Data splits.}
VideoMem has 7000 videos in the training set and 1000 in the validation set (MediaEval workshop~\cite{Sweeney2022MediaEval}).
Past works report results on the validation set as the test labels are not publicly available.
Memento10k is split into 7000 videos for train and 1500 each for validation and test.
We provided our model's outputs to the competition organizers and report results on the test set.

\paragraph{Memorability metrics.}
The memorability score associated with each video in the datasets captures the proportion of people in the original experiments who correctly recognized the video.
We evaluate model's predictions relative to ground-truth (GT) memorability scores, using the Spearman rank correlation (RC $\uparrow$).
Following previous works, we also report the mean squared error (MSE $\downarrow$) to measure the gap between GT and predictions. 

\paragraph{Implementation details.}
We break each video into $T$ uniform segments and pick one frame at random from each segment during training - this acts as data augmentation~\cite{zhou2018trn}.
For inference, we take the middle frame of the segment.
$T{=}5$ works well for Memento10k (1.66fps) and $T{=}7$ for VideoMem (1fps).
When not specified otherwise, we train our model with the Adam optimizer~\cite{kingma2015adam}, learning rate $10^{-5}$, and a step scheduler (for VideoMem only) with step size 10 epochs and multiplier 0.5.

\subsection{Video Memorability Prediction}
\label{subsec:ablation}

We begin with model ablation studies for Memento10k.
VideoMem has some challenges with respect to data leakage (\cref{subsec:videomem_challenging}) and results are presented in \cref{subsec:videomem_ablations}.

\begin{table}[t]
\centering
\small
\tabcolsep=0.10cm
% \resizebox{\linewidth}{!}{% Adjusts the table to line width
\begin{tabular}{l c cc c c  cc}
\toprule
& &  \multicolumn{2}{c}{Embedding} & & & \multicolumn{2}{c}{Memento10k (val)}\\ 
\cmidrule(lr){3-4} \cmidrule(lr){7-8} 
& CLIP & Time & Space & Sampling & Caption & RC $\uparrow$ & MSE $\downarrow$ \\ 
\midrule
\rowcolor{Apricot!30}
1 & ST & F & - & Random & - & \textbf{0.706} & 0.0061 \\
\midrule
2 & T & F & - & Random& - & 0.687 & 0.0062 \\
3 & ST & L & - &Random & - & 0.696 & 0.0059 \\
4 & ST & F & 1D & Random & -  & \textit{0.703} & \textit{0.0057}  \\
5 & ST & F & 2D & Random & - & 0.701 & \textbf{0.0056} \\
6 & ST & F & - & Middle & - & \textit{0.703} & 0.0066 \\
\midrule
% 7 &  ST & F & -  & R & - & 0.706 & 0.0061 & 0.513 & 0.0060 \\ same as R1
\rowcolor{SkyBlue!30}
7 &  ST & F & -  & Random & Orig. & \textbf{0.745} & \textbf{0.0050} \\
% 8 &  ST & F & -  & R & w original captions as input  & 0.745 & 0.0050 & 0.505 & 0.0061  \\
% 3 & w orig captions (train) & 0.6492 & 0.0165 & 0.xxx & 0.xxxx  \\
8 & ST & F & -  & Random & Pred. & \textit{0.710} & \textit{0.0056} \\
% 9 & ST & F & -  & R & w caption prediction & 0.710 & 0.0056 & 0.508 & 0.0061 \\
\bottomrule
\end{tabular}
% }
\vspace{-2mm}
\caption{\textbf{Model ablations.}
Column 1 (C1) compares the impact of using spatio-temporal (ST) features versus temporal (T) features with global average pooling. 
C2 and C3 specify the types of temporal (L: learnable, F: Fourier) and spatial position embeddings used. 
C4 is the frame sampling method used during training. 
C5 indicates whether the video caption (Orig: original caption, Pred: predicted caption) is used in modeling.
\textit{Row 1 (R1) is chosen as the {\color{Apricot}default configuration} for further experiments} and represents the best vision-only model. 
R2-6 evaluate vision model choices: features, position-encodings, and frame sampling methods.
{\color{SkyBlue}R7} presents results with original captions (Orig.) as a part of the model and R8 aims to predict the captions on the fly.
The best results in each section are in \textbf{bold}, with second-best in \textit{italics}.}

\vspace{-4mm}
\label{tab:ablations_memento}
\end{table}

\paragraph{Ablation of vision models.}
\cref{tab:ablations_memento} rows 1-6 show the results of various hyperparameters of the vision model evaluated on the validation set.
Row 1 (R1) achieves best performance and is the \textit{{\color{Apricot}default configuration}} for further experiments.
Using spatio-temporal (ST, R1) image embeddings and not performing global average pooling (R2) shows a small improvement in RC.
Similarly, using Fourier embeddings (R1) is better than learnable ones (R3), perhaps due to the small dataset size.
Surprisingly, using spatial embeddings to identify the $H {\times} W$ tokens reduces performance (R1 \vs~R4 or R5), perhaps due to the pyramidal nature of the CNN representations.
Finally, using random sampling during training (R1) instead of picking the middle frame of the segment (R6) results in a small increase.
In general, the gap between all rows is small, indicating that results are not impacted strongly by hyperparameter changes.
However, spatio-temporal (ST) CLIP embeddings are required to obtain spatio-temporal model attention maps.

\paragraph{Use of captions.}
\cite{Newman2020} introduced captions (descriptions) for the short videos in Memento10k as a way to emphasize semantic categories for predicting memorability.
We modify our model by extending the sequence length of our Transformer encoder to include additional description tokens.
Visual and text tokens are differentiated through a type embedding (additional details in the supplement, \cref{sec:methodExt}).

In \cref{tab:ablations_memento} (bottom) using the original captions (OC) strongly benefits Memento10k as Spearman RC goes up from 0.706 (R1) to 0.745 (R7).
However, when the visual tokens predict both the memorability score and the caption (similar to CLIPCap~\cite{mokady2021clipcap}) the memorability score shows modest improvement (to 0.710, R8).

\paragraph{SoTA comparison.}
Comparison to state-of-the-art works on Memento10k with different setups (val or test split, {\color{SkyBlue}with} / {\color{Apricot}without} captions) is presented in \cref{tab:SOTA}.
Note, our goal is to understand the attentional factors driving video memorability through a model that provides spatio-temporal attention.
Nevertheless, our model with a single feature encoder (CLIP) achieves results comparable to SoTA (Memento10k: 0.706 val, 0.662 test).
With captions, we obtain 0.713 (test).
To interpret model performance reported as RC scores, we note that a model that performs well is expected to approach a human-human consistency RC of 0.73 for \emph{Memento10K}~\cite{Newman2020}.

\begin{table}[t]
\centering
\small
\tabcolsep=0.08cm
% \resizebox{\linewidth}{!}{% Adjusts the table to line width
\begin{tabular}{l c cc cc}
\toprule
& & \multicolumn{4}{c}{Memento10k} \\ 
& & \multicolumn{2}{c}{Test} & \multicolumn{2}{c}{Val}\\
\cmidrule(lr){3-4} \cmidrule(lr){5-6} 
Methods & Caption & RC & MSE & RC & MSE \\ 
% \midrule
% Human~\cite{Newman2020} & - & - & 0.73 & - \\
\midrule
SemanticMemNet \tablevenue{ECCV20} & No & 0.659 & - & - & - \\
M3-S \tablevenue{CVPR23} & No & - & - & 0.670 & 0.0062  \\
\rowcolor{Apricot!30}
Ours (R1 \cref{tab:ablations_memento}) & No & \textbf{0.662} & 0.0065 & \textbf{0.706}  & 0.0061 \\
\midrule
SemanticMemNet \tablevenue{ECCV20} & Yes & 0.663 & - & - & - \\
Sharingan \tablevenue{arXiv} & Yes & - & - & 0.72 & - \\
\rowcolor{SkyBlue!30}
Ours (R7 \cref{tab:ablations_memento}) & Yes & \textbf{0.713} & 0.0050 &\textbf{0.745} & 0.0050 \\
\bottomrule
\end{tabular}
% }
\vspace{-2mm}
\caption{Comparison against SoTA for video memorability. 
Baselines considered are  SemanticMemNet~\cite{Newman2020}, M3-S~\cite{Dumont2023}, and Sharingan~\cite{Harini2023}.
Split-half human-human consistency RC for Memento10k is 0.73.
See supplement \cref{tab:SOTA_combined} for VideoMem.}
\label{tab:SOTA}
% \EK{abobe, report on val?, videomem gives 0.524 on segmented last (should i put that in the SOTA table then?)}}
\vspace{-5mm}
\end{table}

% \begin{table}[t]
% \centering
% \small
% \tabcolsep=0.12cm
% % \resizebox{\linewidth}{!}{% Adjusts the table to line width
% \begin{tabular}{l c c cc}
% \toprule
% & & & \multicolumn{2}{c}{Memento10k}\\ 
% % \cmidrule(lr){4-5} 
% Methods & Caption & Split & RC & MSE \\ 
% \midrule
% Human~\cite{Newman2020} & - & - & 0.73 & - \\
% \midrule
% SemanticMemNet \tablevenue{ECCV20} & No & Test & 0.659 & - \\
% \rowcolor{Apricot!30}
% Ours (R1 \cref{tab:ablations_memento} on test) & No & Test & \textbf{0.662} & 0.0065  \\
% \midrule
% M3-S \tablevenue{CVPR23} & No & Val & 0.670 & 0.0062  \\
% \rowcolor{Apricot!30}
% Ours (R1 \cref{tab:ablations_memento}) & No & Val & \textbf{0.706}  & 0.0061\\
% \midrule
% SemanticMemNet \tablevenue{ECCV20} & Yes & Test & 0.663 & -  \\
% \rowcolor{SkyBlue!30}
% Ours (R7 \cref{tab:ablations_memento} on test) & Yes & Test & \textbf{0.713} & 0.0050 \\
% \midrule
% Sharingan \tablevenue{arXiv} & Yes & Val & 0.72 & - \\
% \rowcolor{SkyBlue!30}
% Ours (R7 \cref{tab:ablations_memento}) & Yes & Val & \textbf{0.745} & 0.0050 \\
% % Ours (PT LaMem) & No & 0.665  & 0.0060 & 0.505  & 0.0059  \\
% \bottomrule
% \end{tabular}
% % }
% \vspace{-2mm}
% \caption{Comparison against SoTA for video memorability. 
% Baselines considered are  SemanticMemNet~\cite{Newman2020}, M3-S~\cite{Dumont2023}, and Sharingan~\cite{Harini2023}. (See supplement \cref{tab:SOTA_combined} for VideoMem).} 

% \label{tab:SOTA}
% % \EK{abobe, report on val?, videomem gives 0.524 on segmented last (should i put that in the SOTA table then?)}}
% \vspace{-5mm}
% \end{table}

Furthermore, our model is trained only on the Memento10k training set, while all baselines train on a combination of image and video memorability datasets.
For example, pretraining on LaMem~\cite{squalli2018Lamem} and fine-tuning on Memento10k improves performance from 0.706 to 0.715.
For completeness, we present cross-domain transfer results of pretraining and fine-tuning our model on image or video memorability datasets and evaluation on all in the supplement, \cref{sec:imagetovideo}.

All further analyses and experiments are conducted using the vision-only model, without incorporating captions.

\begin{figure*}[t]
\centering
\includegraphics[width=\textwidth]{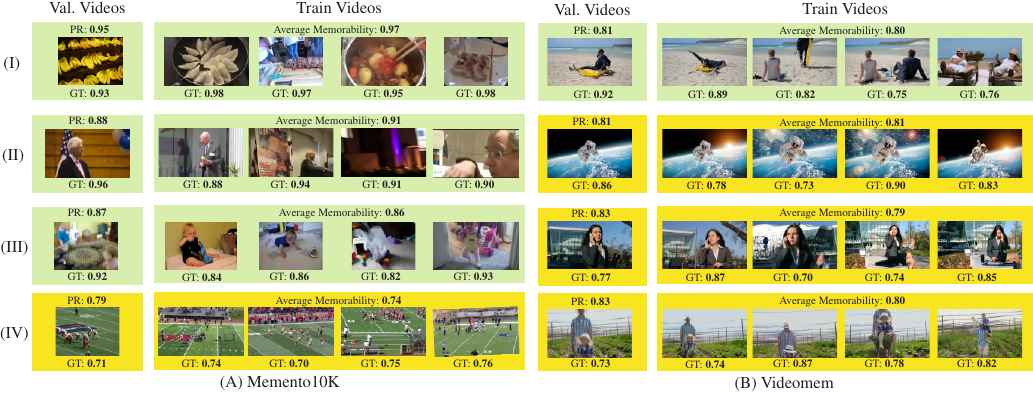}
\vspace{-6mm}
\caption{Nearest neighbor (NN) analysis for videos from Memento10K (left) and VideoMem (right).
We illustrate four validation set videos and for each, four NN from the training set.
We provide the GT memorability score (below), the predicted score on the val set (above), and the average of 4 NN scores from the training set.
In B (right), multiple video clips with high visual similarity between train and validation sets are highlighted with a \emph{yellow} background.
Conversely, the green rows highlight clips that have similar content, but are likely from different source videos.
We discuss how data leakage and variance in GT scores may adversely affect evaluation in \cref{subsec:videomem_challenging}.
}
\label{fig:videomem}
\vspace{-4mm}
\end{figure*}

\subsection{Why is VideoMem challenging?}
\label{subsec:videomem_challenging}
The RC scores on VideoMem~\cite{Cohendet2018} are significantly lower than on Memento10k, even with additional information like captions providing no improvement.
Detailed results can be found in supplement \cref{subsec:videomem_ablations}. 
In fact, most methods achieve RC greater than the human-human RC at 0.481, indicating that models have probably overfit to the dataset, especially as a held-out test set is not available.
As evidence, the code repository of a recent work, M3-S~\cite{Dumont2023}\footnote{\url{https://github.com/theodumont/modular-memorability}} shows that achieving a Spearman RC of 0.5158 is possible after using highly specific random seeds and hyperparameters.

\paragraph{Similar videos across splits.}
We propose a nearest-neighbors (NN) analysis of representations and observe that improving results on VideoMem is challenging due to problems in \emph{split creation}.
We visualize the NN in the training set for each validation video based on $\hat{\bh}_\CLS$, the representation before the MLP regressor.
On VideoMem, from a random sample of 30 validation videos, 14 clips have visually identical NN in the training set.
In contrast, on Memento10k, we are only able to find 1 clip among 30.

\cref{fig:videomem} displays a few videos illustrating this problem.
On Memento10k (left), we see that NNs show semantic awareness and matching (I food, II speaker, IV sports field).
On the other hand, on VideoMem, the NN are (probably) from the same long video.
See right: I surfer, II astronaut, III news anchor, IV farmer.
Given the identical visual stimuli, the model can do no better than predict the average memorability score of the NNs on the training set (which it does).
\Eg~in row II with the astronaut, PR=0.81 is equal to the average memorability, but is away from GT=0.86.
In row IV farmer, PR=0.83 is close to the average 0.80, but away from GT=0.73.
While using multiple feature backbones may help, this is not a satisfactory solution to a fundamental issue of data leakage across splits.
To address this, we attempted to recreate the splits.
However, as the original source video ids are unavailable, it is not easy to detect which video clips belong to the source video.

\paragraph{Implications for data collection.}
We encourage researchers to analyze new datasets before they are released.
Information about the video source and split creation process are crucial aspects for any dataset.
Additionally, memorability scores are a measure of consensus among viewers and are therefore closely tied to the number of viewers per video.
While LaMem averages 80 scores per image,
Memento10K has over 90 annotations per video,
Videomem averages 38 annotations per video, much smaller than the others.
This variance in GT scores is also observed in \cref{fig:videomem} (B-II), videos of the same astronaut have GT scores varying from 0.73 to 0.90, making learning difficult.

\subsection{Comparing Model Attention and Human Gaze}
\label{sec:attVGazeText}

\paragraph{Setup.}
To compare the human gaze fixation density maps and model-generated attention maps,
we first min-max normalize them to $[0, 1]$.
Next, we compute multiple popular metrics
\footnote{We compute all metrics following the methods used by \url{https://github.com/imatge-upc/saliency-2019-SalBCE/blob/master/src/evaluation/metrics_functions.py}}
in saliency evaluation~\cite{bylinskii2018different}:
AUC-Judd~\cite{judd2009learning},
Normalized Scanpath Saliency (NSS)~\cite{borji2012quantitative},
Linear Correlation Coefficient (CC)~\cite{Ouerhani2004CC}, and
Kullback-Leibler Divergence (KLD)~\cite{Tatler2005KLD, Rajashekar2004KLD}.

We split participants into two random groups and for a given video, compute agreement between the two groups using the saliency metrics.
These human-human (H-H) agreement scores are averaged over 10 random split iterations and then across videos.
H-H scores act as a ceiling against which our model-human (M-H) agreement scores are compared.
To obtain chance-level performance, we compute H-H agreement scores but now with shuffled videos (H-H Shuff.).

\newcommand{\meansemtwo}[2]{\begin{tabular}{@{}c@{}} #1 \\[-0.3em] {\scriptsize${\pm}$#2} \end{tabular}}

\begin{table}[t]
\centering
\small
\tabcolsep=0.10cm
\begin{tabular}{l c c c c c c}
\toprule
& \multicolumn{3}{c}{Memento10k} & \multicolumn{3}{c}{VideoMem} \\ 
\cmidrule(lr){2-4} \cmidrule(lr){5-7} 
Metrics & M-H & H-H & H-H Shuff. & M-H & H-H & H-H Shuff.\\ 
\midrule
AUC-J $\uparrow$ & \meansemtwo{0.89}{0.007} & \meansemtwo{0.90}{0.001} & \meansemtwo{0.70}{0.002} & \meansemtwo{0.89}{0.007} & \meansemtwo{0.80}{0.002} & \meansemtwo{0.55}{0.001} \\
% \midrule
AUC-P $\uparrow$ & \meansemtwo{82.91}{1.65} & - & - & \meansemtwo{88.88}{1.29} & - & - \\
% \midrule
NSS $\uparrow$ & \meansemtwo{1.95}{0.074} & \meansemtwo{3.07}{0.024} & \meansemtwo{0.84}{0.022} & \meansemtwo{2.00}{0.068} & \meansemtwo{3.12}{0.023} & \meansemtwo{0.23}{0.012} \\
% \midrule
CC $\uparrow$ & \meansemtwo{0.46}{0.014} & \meansemtwo{0.49}{0.003} & \meansemtwo{0.16}{0.003} & \meansemtwo{0.27}{0.007} & \meansemtwo{0.27}{0.018} & \meansemtwo{0.03}{0.001} \\
% \midrule
KLD $\downarrow$ & \meansemtwo{1.48}{0.035} & \meansemtwo{2.17}{0.023} & \meansemtwo{4.61}{0.022} & \meansemtwo{2.65}{0.020} & \meansemtwo{4.02}{0.018} & \meansemtwo{6.49}{0.013} \\
\bottomrule
\end{tabular}
\vspace{-2mm}
\caption{Comparing gaze fixation maps against model's attention map via different metrics, along with human-human split-half reliability scores over 10 iterations. $\uparrow$ ($\downarrow$) indicates higher (lower) is better. 
M-H: Model-human; H-H: Human-human; and H-H Shuff.: Human-Human\_shuffled (random performance).}
\label{tab:fixation_vs_attention}
\vspace{-5mm}
\end{table}

\begin{figure*}[t]
\centering
\includegraphics[valign=m,width=0.8\textwidth]{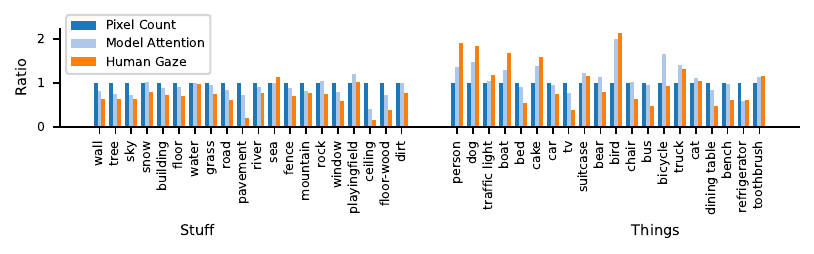}
\raisebox{0.55cm}{\includegraphics[valign=m,width=0.19\textwidth]{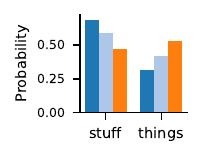}}
\vspace{-3mm}
\caption{
Analysis of panoptic segmentation for the most common 40 classes (20 stuff, 20 things).
\underline{Left} shows normalized pixel counts (blue), model attention-weighted counts (light blue), and human gaze-weighted counts (orange).
Both, model and humans, show lower affinity for stuff classes and higher for thing classes, indicating their importance in memorability.
\underline{Right} Pixel counts are accumulated across stuff and thing classes, highlighting the above trend clearly.
Best viewed on screen with zoom.
}
\vspace{-2mm}
\label{fig:panoptic}
\end{figure*}

\begin{figure}[t]
\centering
\includegraphics[width=\linewidth]{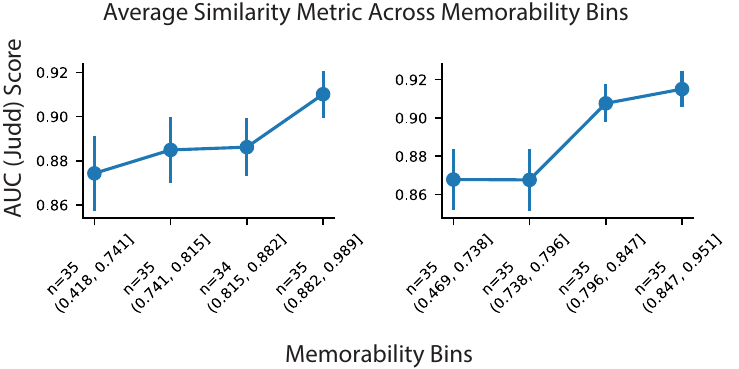}
\vspace{-3mm}
\caption{Gaze \vs~attention similarity metrics with AUC-Judd scores on the Y-axis and Ground Truth on the X-Axis. (See supplement~\cref{subsec:metrics_and_complexity}, \cref{fig:metrics} for other metrics and their trends.) \underline{Left}: Memento10k,
\underline{Right}: VideoMem. Error bars depict SEMs.
}
\label{fig:true_bins}
\vspace{-3mm}
\end{figure}

\paragraph{Results.}
While \cref{fig:teaser} shows qualitative results of human gaze and model attention,
\cref{tab:fixation_vs_attention} indicates that there is a high degree of M-H similarity across both datasets.
We observe that metrics (AUC-J, CC) often approach the H-H scores, and importantly, significantly improve over random chance (H-H Shuff.).
In \cref{fig:true_bins}, we plot AUC-Judd and NSS against GT memorability bins and observe that the similarity between model attention and human gaze maps increases with GT memorability scores in both datasets.
This suggests that highly memorable videos have clear regions of focus for both humans and the model.
Please refer to the supplement~\cref{subsec:metrics_and_complexity} for other metrics.

Furthermore, we replicate these results on image datasets by using a model pretrained on LaMem~\cite{khosla_understanding_2015} and fine-tuned on FIGRIM~\cite{figrim} (supplement~\cref{subsec:image_memorability}).

\paragraph{Center bias.}
Among metrics, we also considered the shuffled AUC (sAUC)~\cite{zhang2008sun}, but it tends to unjustly penalize valid central predictions~\cite{Neil2020}.
Therefore, we introduce a metric to measure relative similarity, \emph{AUC-Percentile}.
For a given video, we compare the true AUC-Judd between model attention and human gaze against a distribution of AUC-Judd values calculated by comparing model attention from that video and human gaze from other randomly selected videos.
The percentile of the true AUC-Judd score within the distribution of random AUC-Judd scores estimates the probability that the true score is video-specific and is not obtained by chance or due to center bias. 
For instance, a model driven purely by center-bias (using a 2D Gaussian, $\sigma{=}$10\% of the scene height~\cite{Lyu2020AttCorr}) yields an average AUC-Percentile score of \meansem{76.17}{2.62} on Memento10K and \meansem{68.47}{2.82} for VideoMem.
Results in \cref{tab:fixation_vs_attention} show that our model's AUC-P scores at \meansem{82.91}{1.65} and \meansem{88.88}{1.29} exceed these center-bias-driven AUC-P scores.

Another approach to rule out the possibility that the high M-H similarity is due to center bias involves a direct comparison between the performance of the previously explained Gaussian-based center bias model~\cite{Lyu2020AttCorr} and our proposed gaze prediction model. We use the Gaussian to simulate central fixation and calculate median AUC-J score across frames per video.
Compared to the Gaussian, our model is better aligned with human fixations across videos on both datasets, Memento10K ($p=0.003$) and VideoMem ($p=5.80\times 10^{-12}$).

\subsection{Panoptic Segmentation}

We extract panoptic segmentation labels from MaskFormer~\cite{maskformer}, a SoTA model for segmentation, on the $T$ selected video frames (see supplement \cref{fig:panoptic_all} for examples).
We use the COCO-stuff hierarchy~\cite{Caesar2018coco} to classify labels as \emph{stuff} or \emph{things}.
We create three sets of counts:
(i)~\emph{Pixel Count} sums the number of pixels attributed to each label across frames and videos (normalized by the total number of pixels in the frame).
(ii)~\emph{Model Attention} weighted counts multiply the attention map with segmentation masks of each category, summing across frames and videos.
(iii)~\emph{Human Gaze} weighted counts are similar and multiply gaze fixation densities with segmentation masks.

\paragraph{Stuff \vs things classes.}
We consider the most prevalent \emph{stuff} and \emph{things} labels (20 each) across the 140 videos of the eye-tracking dataset and observe that attention increases/decreases relative to normalized pixel counts in similar ways for models and humans (\cref{fig:panoptic} left).
Specifically, we observe a tendency for decreased attention to \emph{stuff} and increased attention to \emph{things}, which is clear in the cumulative distributions (\cref{fig:panoptic} right).

\paragraph{Simple \vs~complex videos.}
Panoptic segmentation also allows us to answer a crucial question about the impact of video complexity on model-human alignment.
We split our videos into simple and complex based on the number of objects averaged over frames (median split).
Comparing model-human and human-human alignment in these videos, we find no significant differences in most metrics (see supplement \cref{subsec:metrics_and_complexity}) suggesting that our results are not influenced by the complexity of videos.

\subsection{Temporal Attention}

\begin{figure}[t]
\centering
\includegraphics[width=\linewidth]{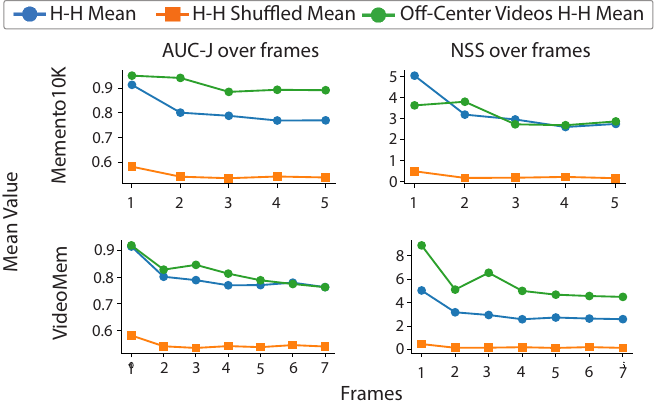}
\vspace{-2mm}
\caption{Framewise split-half AUC-J and NSS scores for Memento10K (left) and VideoMem (right).
The x-axis shows sub-sampled frames at $T{=}5$ for Memento10K and $T{=}7$ for VideoMem.
The blue line (H-H) indicates the framewise alignment between gaze patterns, averaged over all 140 videos. 
The green line captures framewise alignment averaged over 35/140 videos that have most off-center saliency in the initial frames.
The orange line represents H-H shuffled, mean alignment when gaze patterns are compared across random videos.}
\label{fig:framewise_split}
\vspace{-2mm}
\end{figure}

We first analyze whether humans look at similar regions across frames of a video and find that they are more consistent in the initial frames of the video as compared to later frames, see \cref{fig:framewise_split} (blue).
However, it is possible that this result is driven by center bias if most videos have salient central regions at the start.
To rule this out, we identify a subset of videos that have off-center salient regions in the initial frames.
\footnote{We adopt DeepGaze~\cite{Kummerer2017DeepGaze}
and compute saliency maps for $T$ video frames.
Next, we compute a distance between the predicted saliency map and a center bias, modeled as a Gaussian, and sort the videos in decreasing distance.
For this analysis, we consider $25^\text{th}$ percentile most off-centered videos for Memento10k and VideoMem separately.}
\cref{fig:framewise_split} (green) shows us that there is stronger consensus across participants for the off-centered videos, and this too goes down as the video progresses.

\begin{figure}[t]
\centering
\includegraphics[width=\linewidth]{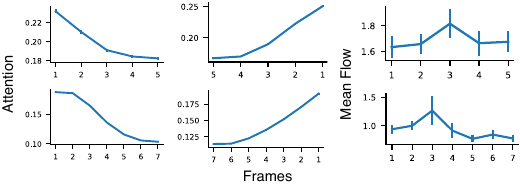}
\vspace{-3mm}
\caption{
\underline{Left}: Distribution of temporal attention across video frames in normal order, showing peak at the early frames.
\underline{Middle}: Distribution of temporal attention across video frames in \emph{reversed} order as a control to rule out position bias.
\underline{Right}: Mean optical flow magnitude across frames to rule out motion as a bias for the stronger temporal attention at the beginning.
The x-axis indicates the number of sub-sampled frames; $T{=}5$ for Memento10K (top) and $T{=}7$ for VideoMem (bottom). 
}
\label{fig:temporal}
\vspace{-3mm}
\end{figure}

Next, to ascertain whether our model displays similar temporal patterns of attention, we compute attention scores as $\alpha \in \bbR^{T \times HW}$ and sum over the spatial dimensions to obtain temporal attention, $\alpha_T \in \bbR^{T}$. 
As visualized in \cref{fig:temporal} left, our model preferentially attends to the initial frames of the video sequence, without any architectural bias towards this.
We further rule out two possibilities:
(i)~reversing the frames (and preserving the same temporal position embeddings), we observe that the model still gives more attention to early frames (now appearing at the end, \cref{fig:temporal} middle);
(ii)~computing optical flow magnitude~\cite{raft} per frame, averaged across all pixels, we find that motion is strongest around the middle (\cref{fig:temporal} right) and cannot be the reason for increased attention to early frames.

Therefore, we conclude that our model, only trained to predict memorability scores, has learned to attend to the visual information that most participants look at earlier on in the videos.

\section{Conclusion}
\label{sec:conclusion}
We adopted a simple CNN+Transformer model that not only matches SoTA in predicting video memorability scores, but also enables exploring the underlying spatio-temporal attention mechanisms.
Furthermore, we collected human gaze data to compare against model attention and observed that the model and humans look at similar regions.
We also discovered novel semantic attention patterns relevant for video memorability.
On the temporal dimension, the model exhibited strong preference for early frames of the videos, mimicking temporal patterns in human attention.
We also analyzed a widely used video memorability dataset, identifying several critical issues that researchers must consider when constructing new datasets.

\paragraph{Limitations.}
The current datasets have 10k videos each.
A model trained on them may not generalize well to any video from the internet, especially in specific domains where the visual stimuli are typically similar across all clips, \eg~identifying memorable parts from a lecture video.
Additionally, the model processes extracted frames rather than full videos, which may result in the loss of important details for memorability and could affect comparison with human data, where viewers see the entire video.

\paragraph{Acknowledgments.}
We thank Sriya Ravula and Dr. Priyanka Srivastava for help with eye-tracking equipment and its setup.
The study was supported by the IIIT-H Faculty Seed Fund (VS) and an Adobe Research Gift (MT).

%%%%%%%%% REFERENCES
{\small
\bibliographystyle{ieee_fullname}
\bibliography{main}
}

\clearpage

\maketitlesupplementary
\appendix

In the supplementary material, we present an expanded set of results and analyses to better understand of our work.
\cref{sec:eyetracking} provides details of our eye-tracking setup, methodology, and apparatus. 
\cref{sec:imagetovideo} examines the performance of our model on image memorability tasks and impact of transfer learning.
\cref{sec:qualitative} presents additional experiments and results:
(i)~model ablation results and comparison to state-of-the-art on VideoMem~\cite{Cohendet2018};
(ii)~detailed qualitative analysis on both Memento10K~\cite{Newman2020} and VideoMem datasets including human gaze and model attention maps;
(iii)~additional similarity metrics and assessment of the impact of video complexity; and
(iv)~results comparing human gaze \vs~model attention on the FIGRIM~\cite{figrim} image memorability dataset.
\cref{sec:methodExt} explains the integration of text captions into our model, and the corresponding results.
Finally, \cref{sec:panoptic_supp} discusses the results of applying panoptic segmentation to better understand the semantic concepts in the scene.

\section{Eye-tracking Setup}
\label{sec:eyetracking}

\subsection{Experiment Setup Details}
\label{sec:eyetracking_setup}
The eye-tracking experiment is structured in the form of a continuous recognition experiment, where we present participants with a series of videos and instruct them to press the \texttt{SPACEBAR} when they recognize a video as being a repeat of one they had seen earlier in the sequence. As feedback for participants, we change the background color of the display to \texttt{GREEN} in case of a true positive and \texttt{RED} in case of a false positive.

We recruit $20$ participants to watch $200$ videos each from the Memento10K and VideoMem datasets, each participant watching videos exclusively from one dataset.

We select participants based on a strict criterion relating to their visual acuity, only considering individuals with a refractive error (eyeglass power) within the range of $[-1, +1]$ diopters. We establish this criterion in order to maintain a standard level of natural visual acuity among participants. Additionally, we require all participants to view the videos without the aid of eyeglasses, ensuring that any corrective lenses did not affect the pupil tracking device.

For the participants watching videos from Memento10K we display videos in their original size and aspect ratio on a screen of size $1024{\times} 768$. For participants watching videos from VideoMem, we display videos in their original aspect ratio, resized to fit the screen width. For example, we convert videos with size $1920 {\times} 1080$ to $1024 {\times} 576$, maintaining the aspect ratio of $1.77$.

We calibrate and validate pupil positions after every $20$ videos for Memento10K and 10 videos for VideoMem (approximately $1$ minute). Participants use a mounted chin-rest while viewing videos, placed at a distance of \SI{35}{\centi\metre} from the screen.

The primary interest is in capturing the participants' fixations while engaged in a memory game similar to the original studies of Memento10K and VideoMem.

\begin{figure}[t]
\centering
\includegraphics{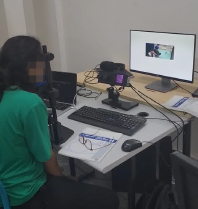}
\caption{Participant watching videos from Memento10K during the eye-tracking experiment (face anonymized).}
\label{fig:eyetracking_participant}
\end{figure}

The eye-tracking study involving human participants was reviewed and approved by the Institute Review Board (IRB). The participants provided their written informed consent to participate in the study.

\begin{figure*}[t]
\centering
\includegraphics[width=\textwidth]{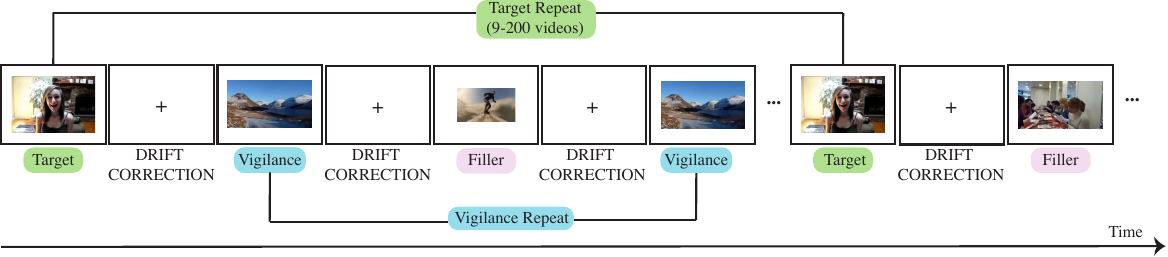}
\caption{Design of eye-tracking experiment.
A subject watches alternating videos and drift correction fixation crosses (typically between \SI{0.5}{\second} to \SI{1}{\second}).
A vigilance video (one of 40) is repeated in a short interval of 2-3 videos to ensure that the subject is alert, while the target videos (one of 20) have a lag of at least 9 videos.
Filler videos (80) are not repeated.}
\label{fig:continuous_recognition}
\end{figure*}

\subsection{Eye-tracking Procedure}
\label{sec: eyetracking_procedure}
The main procedure of the experiment (sequence in which videos are shown) is presented in \cref{fig:continuous_recognition}.
An instance of a participant watching the videos can be seen in \cref{fig:eyetracking_participant}.

\paragraph{Video selection.}
We select $200$ videos each from the validation sets consisting of $1500$ videos in Memento10K and $1000$ in VideoMem. To ensure a representative and varied selection of videos, we use a two-step process:

\begin{enumerate}
\item \textbf{Clustering:} We initially cluster videos based on their visual features. We extract the average CLIP ResNet~\cite{radford2021clip} embeddings from selected frames of each video — $T{=}5$ linearly spaced frames for videos from the Memento dataset and $T{=}7$ linearly spaced frames from Videomem.
We then group these videos into $28$ distinct clusters using K-Means Clustering, providing a structured framework for subsequent selection.
We choose $K{=}28$ by visually inspecting the quality of clusters (generated from hierarchical clustering) for values around $30$.

\item \textbf{Binning:}  Following clustering, we bin videos based on their ground truth memorability scores, creating 10 distinct bins. This stratification allows for a balanced representation of memorability levels within the selected videos.
\end{enumerate}

We select videos for the experiment through the following sampling strategy:
Initially, we sample one video from each cluster-bin combination, ensuring broad coverage across all memorability levels and visual characteristics. In instances where the initial sampling does not yield $200$ videos, we sample for a second iteration. This round involves selecting an additional video from some cluster-bin combinations, again governed by the availability of videos within each category. To adhere to the desired total of $200$ videos, we uniformly remove any excess videos from the sampled pool. We randomly select and remove these excess videos from the cluster-bin combinations, ensuring an even distribution across all categories.

From these selected $200$ videos, the experiment design requires a refined set of $140$ unique videos ($20$ target repeats, $40$ vigilance repeats, and $80$ fillers). We, therefore, randomly select videos for each category (vigilance, target, and regular) from the pool of $200$ videos, ensuring that each category had a distinct set of videos. The target and vigilance repeats are the same across all participants. For each experimental run, we use a unique order of video presentations. This involves mixing regular videos with the vigilance and target videos and then randomly shuffling this combined set. We constrain the placement of repeated vigilance videos to a lag of $2-3$ videos, while for target videos, we maintain a minimum lag of $9$ videos (similar to VideoMem and Memento10k).

\subsection{Details of Gaussian blur}
\label{sec:gaussian_blur}
To account for the visual field of a participant, we apply a Gaussian blur to fixation maps obtained from the experiment. The standard deviation ($\sigma$) of the Gaussian blur is calculated using the formula:
\begin{equation}
\sigma = \frac{\text{Pixels Per Degree}}{2.355} \, .
\end{equation}
Here, $2.355$ is a constant derived from the assumption that the visual angle corresponds to the Full Width at Half Maximum (FWHM).
The Pixels Per Degree (\text{PPD}) is computed as follows:
\begin{equation}
\text{PPD} = \frac{2 \times d \times \tan(\frac{\theta}{2})}{h \times y} \, ,
\end{equation}
where, $d$ is the distance of the participant to the screen (\SI{13.77}{inch} or \SI{35}{\centi\metre}),
$\theta$ is the visual angle (assumed to be 1\textdegree), 
$h$ is the height of the screen (\SI{23.5}{inch}), and
$y$ is the vertical resolution of the screen ($768$ pixels in our case).

\subsection{Metric: AUC-Percentile}
For a video $V_k$ and video frame $f_{ik}$, the true frame similarity score is computed using AUC-Judd between the model's attention map $\alpha_{ik}$ and the corresponding gaze fixation density map $G_{ik}$. Then, for each video, we compute a true video similarity score by averaging the  true frame similarity scores.

We perform a permutation test by comparing the attention map $\alpha_{ik}$ of frame $f_{ik}$ against the fixation map $G_{il}$ from frame $f_{il}$ of a randomly chosen different video $V_l, l\neq k$.

This process is repeated 100 times, yielding a distribution of 100 video-level similarity AUC-Judd scores under the null hypothesis of no specific relationship between model attention and human fixations.
% $\alpha_{ik}$ and $G_{il}$. 

We denote AUC-Percentile for each video as the percentile of the true video similarity score within the distribution of permuted AUC-Judd scores. A high AUC-Percentile indicates a strong alignment between the model's attention map and the human gaze fixation density map for the same video, relative to a null distribution of comparisons between different videos. For example, an AUC-Percentile of $80$ implies that there is a less than $20\%$ chance that the observed alignment between the model's attention map and the human gaze fixation density map could be attributed to chance or general center-bias in the data.

% \begin{table}[h]
% \centering
% \resizebox{\linewidth}{!}{% Adjusts the table to line width
% \begin{tabular}{@{}lllllll@{}}
% \toprule
% \multirow{2}{*}[-1em]{Train on} & \multicolumn{2}{c}{Lamem} & \multicolumn{2}{c}{Memento} & \multicolumn{2}{c}{Videomem} \\ 
% \cmidrule(lr){2-3} \cmidrule(lr){4-5} \cmidrule(lr){6-7}
% & RC & MSE & RC & MSE & RC & MSE\\ 
% \midrule
% % No text & \multicolumn{4}{c}{Table 1 row 1} \\
% Lamem & 0.720 & 0.0073 & 0.525 & 0.0119 & 0.388 & 0.013 \\
% \addlinespace
% Memento10k & 0.526 & 0.0024 & 0.706 & 0.0061 & 0.441 & 0.015 \\
% \addlinespace

% Videomem & 0.522 & 0.0130 & 0.524 & 0.0089 & 0.513 & 0.0060 \\
% \addlinespace

% \begin{tabular}[c]{@{}l@{}}PT Lamem, \\ FT Memento10k\end{tabular} & 0.627 & 0.0183 & 0.715 &0.0550 & 0.446 & 0.0118 \\
% \addlinespace 

% \begin{tabular}[c]{@{}l@{}}PT Lamem, \\ FT Videomem\end{tabular} & 0.610 & 0.0105 & 0.527 & 0.0095 & 0.505 & 0.0059 \\
% \addlinespace

% \begin{tabular}[c]{@{}l@{}}PT Lamem, \\ FT Memento10k\\+Videomem\end{tabular} & 0.591 & 0.0135 & 0.692 & 0.0058 & 0.499 & 0.0064 \\
% \bottomrule
% \end{tabular}
% }
% \caption{Image to video transfer. image memorability models can be  adapted or downstreamed for  video memorability.
% }
% \label{tab:image_to_video}
% \end{table}

\begin{table*}[t]
\small
\tabcolsep=0.10cm
\centering
% \resizebox{\linewidth}{!}{% Adjusts the table to line width
\caption{Results of transferring an image/video memorability model to images/videos.
Datasets:
LM: LaMem~\cite{khosla_understanding_2015},
M10k: Memento10k~\cite{Newman2020},
VM: VideoMem~\cite{Cohendet2018}, and 
FG: FIGRIM~\cite{figrim}.
Training strategy: P for pretraining and F for fine-tuning. Results reported on validation set.}
\label{tab:image_to_video}
\begin{tabular}{l cccc cc cc cc cc}
\toprule
& \multicolumn{4}{c}{Train on} & \multicolumn{2}{c}{LaMem} & \multicolumn{2}{c}{Memento10k} & \multicolumn{2}{c}{VideoMem}  & \multicolumn{2}{c}{FIGRIM} \\ 
\cmidrule(lr){2-5} \cmidrule(lr){6-7} \cmidrule(lr){8-9} \cmidrule(lr){10-11}  \cmidrule(lr){12-13}
& LM & M10k & VM & FG & RC & MSE & RC & MSE & RC & MSE  & RC & MSE\\ 
\midrule
% No text & \multicolumn{4}{c}{Table 1 row 1} \\
1 & F & - & - & - & \textbf{0.729} & 0.0074 & 0.526 & 0.0220 & 0.382 & 0.0233 & 0.647 & 0.0168 \\
2 & - & F & - & - &  0.547 & 0.0273 & \textit{0.706} & 0.0061 & 0.439 & 0.0165  &  0.351 & 0.0525 \\
3 & - & - & F & - &  0.549 & 0.0147 & 0.525 & 0.0089 & \textbf{0.513} & 0.0060  & 0.501 &  0.0355 \\
% 4 & - & - & - & F & xx \\
\midrule
4 & P & F & - & - & 0.679 & 0.0161 & \textbf{0.718} &0.0568 & 0.446 & 0.0144   & 0.634 & 0.0318\\
5 & P & - & F & - & 0.688 & 0.0090 & 0.459 & 0.0096 & 0.504 & 0.0059  & 0.627 & 0.0237 \\
6 & P & - & - & F & 0.678 & 0.0113 & 0.507 & 0.0130 & 0.392 & 0.0191  & \textbf{0.742} & 0.0123 \\
7 & P & F & F & - & 0.664 & 0.0135 & 0.689 & 0.0058 & 0.483 & 0.0062  & 0.626 & 0.0273 \\
\bottomrule
\end{tabular}
% }
\end{table*}

\begin{figure*}[t]
\centering
\includegraphics[width=\textwidth]{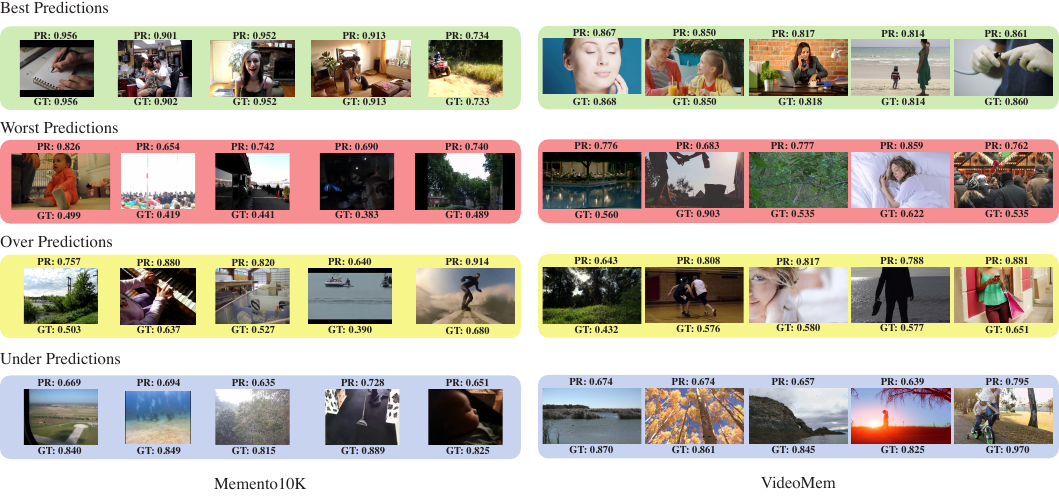}
\caption{Qualitative analysis of different predictions of our model over Memento10K \textit{(Left)} and VideoMem \textit{(Right)}.
Ground-truth (GT) and predicted (PR) memorability scores are annotated at the bottom and top of each frame, representative of the videos.
Best viewed on screen by zooming in.}
\label{fig:qualitative_analysis}
\end{figure*}

\section{Transferring from/to Image Memorability}
\label{sec:imagetovideo}

To ascertain the reliability of our simple approach, we evaluate on image memorability tasks by considering the image as a ``video'' of $T{=}1$.
As seen in \cref{tab:image_to_video},
on the LaMem dataset~\cite{khosla_understanding_2015} we match SoTA results (0.720 RC~\cite{squalli2018Lamem}).
On the FIGRIM dataset~\cite{figrim}, we achieve results close to human performance (0.74 RC~\cite{figrim}). 
Previous studies~\cite{Newman2020} pretrain models on image memorability datasets and then fine-tune them for video memorability prediction.
\cref{tab:image_to_video} R2 \vs~R4 shows a small improvement in Memento10k RC score from 0.706 to 0.718 with LaMem pretraining.
However other results do not improve.
We also observe that training on one dataset and evaluating on another (rows 1-3) usually leads to significant degradation and is an important problem for future work.

\section{Additional Results and Qualitative Analysis}
\label{sec:qualitative}

In this section, we present the results for video memorability prediction on the VideoMem dataset, followed by a qualitative analysis of the model's performance. Finally, we explore the alignment between human gaze and model attention through various analyses on both video and image memorability datasets.

% Next, we compare human gaze and model attention using additional metrics and analyse the impact of video complexity. Finally, we extend our findings to image memorability using the FIGRIM datset.
% C.1 VideoMem results
% - ok trends on ablations; captions don't help
% - we are worse than sota, likely due to 4.2 things
% C.2 Qualitative results
% - Best, worst, over, under.
% - Visualizing gaze and attention maps.
% C.3 Additional results on comparing Human gaze vs model attention
% - More metrics
% - Video complexity
% C.4 Image memorability

\subsection{Video Memorability prediction for Videomem}
\label{subsec:videomem_ablations}

\begin{table*}[t]
\centering
\tabcolsep=0.12cm
% \resizebox{\linewidth}{!}{% Adjusts the table to line width
\begin{tabular}{l c c c c c  cc cc}
\toprule
& & \multicolumn{2}{c}{Embedding} & & & \multicolumn{2}{c}{Memento10k (val)} & \multicolumn{2}{c}{VideoMem (val)} \\ 
\cmidrule(lr){3-4} \cmidrule(lr){7-8} \cmidrule(lr){9-10}
& CLIP & Time & Space & Sampling & Caption & RC $\uparrow$ & MSE $\downarrow$ & RC $\uparrow$ & MSE $\downarrow$ \\ 
\midrule
\rowcolor{Apricot!30}
1 & Spatio-Temporal& Fourier & - & Random & - & \textbf{0.706} & 0.0061 & \textit{0.513} & \textit{0.0060} \\
\midrule
2 & Temporal & Fourier & - & Random& - & 0.687 & 0.0062 & 0.508 & 0.0064 \\
3 & Spatio-Temporal & Learnable & - &Random & - & 0.696 & 0.0059 & 0.502 & 0.0060 \\
4 & Spatio-Temporal & Fourier & 1D & Random & -  & \textit{0.703} & \textit{0.0057} & 0.506 & \textbf{0.0059} \\
5 & Spatio-Temporal & Fourier & 2D & Random & - & 0.701 & \textbf{0.0056} & 0.505 & \textit{0.0060} \\
6 & Spatio-Temporal & Fourier & - & Middle & - & \textit{0.703} & 0.0066 & \textbf{0.515} & \textbf{0.0059} \\
\midrule
% 7 &  ST & F & -  & R & - & 0.706 & 0.0061 & 0.513 & 0.0060 \\ same as R1
\rowcolor{SkyBlue!30}
7 &  Spatio-Temporal & Fourier & -  & Random & Original & \textbf{0.745} & \textbf{0.0050} & 0.505 & 0.0061  \\
% 8 &  ST & F & -  & R & w original captions as input  & 0.745 & 0.0050 & 0.505 & 0.0061  \\
% 3 & w orig captions (train) & 0.6492 & 0.0165 & 0.xxx & 0.xxxx  \\
8 & Spatio-Temporal & Fourier & -  & Random & Predicted & \textit{0.710} & \textit{0.0056} & 0.508 & 0.0061 \\
% 9 & ST & F & -  & R & w caption prediction & 0.710 & 0.0056 & 0.508 & 0.0061 \\
\bottomrule
\end{tabular}
% }
\vspace{-2mm}
\caption{\textbf{Model ablations.}
Column 1 (C1) compares the impact of using spatio-temporal features versus temporal features with global average pooling. 
C2 and C3 specify the types of temporal and spatial position embedding used. 
C4 is the frame sampling method used during training.
C5 indicates whether the video caption is used in modeling.
\textit{Row 1 (R1) is chosen as the default configuration for further experiments} and represents the best vision-only model. 
R2-6 evaluate varying visual choices: features, position-encoding, and frame sampling methods. 
R7 presents results with original captions as a part of the model and R8 aims to predict the captions on the fly.
The best results in each section are in \textbf{bold}, with second-best in \textit{italics}.}

\label{tab:ablations_combined}
\end{table*}

\begin{table*}[t]
\centering
\small
\tabcolsep=0.12cm
% \resizebox{\linewidth}{!}{% Adjusts the table to line width
\begin{tabular}{l c cc cc cc cc}
\toprule
& & \multicolumn{2}{c}{Memento10k (test)} & \multicolumn{2}{c}{VideoMem (test)} & \multicolumn{2}{c}{Memento10k (val)} & \multicolumn{2}{c}{VideoMem (val)} \\ 
\cmidrule(lr){3-4} \cmidrule(lr){5-6} \cmidrule(lr){7-8} \cmidrule(lr){9-10}
Methods & Caption & RC & MSE & RC & MSE & RC & MSE & RC & MSE \\ 
\midrule
% Human & - & - & 0.73 & - & 0.481  & - \\
% \midrule
VideoMem \tablevenue{ICCV19} & No  & - & - & 0.494 & -  & - & - & 0.503 & -  \\
SemanticMemNet \tablevenue{ECCV20} & No  & 0.659 & - & - & - & - & - & 0.555 & - \\
M3-S \tablevenue{CVPR23} & No  & - & - & - & - & 0.670 & 0.0062 & 0.563 & 0.0046 \\
\rowcolor{Apricot!30}
Ours (R1 \cref{tab:ablations_combined}) & No  & 0.662 & 0.0065 & - & - & 0.706  & 0.0061 & 0.513 & 0.0060 \\
\midrule
SemanticMemNet & Yes  & 0.663 & - & - & - & - & - & 0.556 & - \\
Sharingan \tablevenue{arXiv}  & Yes & - & - & - & - & 0.72 & - & 0.6 & - \\
\rowcolor{SkyBlue!30}
Ours (R7 \cref{tab:ablations_combined}) & Yes & 0.713 & 0.0050 & - & - & 0.745 & 0.0050 & 0.505 & 0.0061  \\
\bottomrule
\end{tabular}
\vspace{-2mm}
\caption{Comparison against SoTA for video memorability on both test and validation sets for Memento10k and VideoMem.
Baselines considered are VideoMem~\cite{Cohendet2018}, SemanticMemNet~\cite{Newman2020}, M3-S~\cite{Dumont2023}, and Sharingan~\cite{Harini2023}.
Human-human split-half consistency scores are 0.73 for Memento10k and 0.481 for VideoMem.
}
\label{tab:SOTA_combined}
\vspace{-5mm}
\end{table*}

Expanding on the model ablations for Memento10k in \cref{subsec:ablation} (of the main paper), \cref{tab:ablations_combined}  shows results for VideoMem, which generally follows similar trends, with Row 1 (R1) achieving the best results.
However, random sampling during training does not improve performance and including or predicting captions has no impact, perhaps due to the noise in the captions.

SoTA comparisons are shown in \cref{tab:SOTA_combined}.
As the test set memorability scores (labels) for VideoMem are not available, no previous work apart from the creators of the dataset have evaluated on a held-out test set.
Instead, all approaches likely overfit on the validation set with RC scores much higher than the human-human consistency RC at 0.481.
Our scores are lower than other SoTA methods, likely due to the challenges discussed in \cref{subsec:videomem_challenging}.
However, we suspect that other models that leverage multiple modalities are strongly overfitting on this dataset.

\subsection{Qualitative Analysis}
\label{subsec:qualitative}

We provide a qualitative analysis of the model's predictions and the alignment of its attention maps with human gaze, highlighting the model's successes and failures.

\paragraph{Best, worst, over, and under predictions.}
A few qualitative examples of different predictions of our model across both datasets can be seen in \cref{fig:qualitative_analysis}.
The model seems to perform well on videos with a clear subject (face, a man playing with their dog, \etc).
Worst predictions (over and under) are observed on underexposed (dark) videos.
The model tends to over-predict on certain videos with clutter, while under-predict on scenic videos.

\begin{figure*}[t]
\centering
\includegraphics[width=\textwidth]{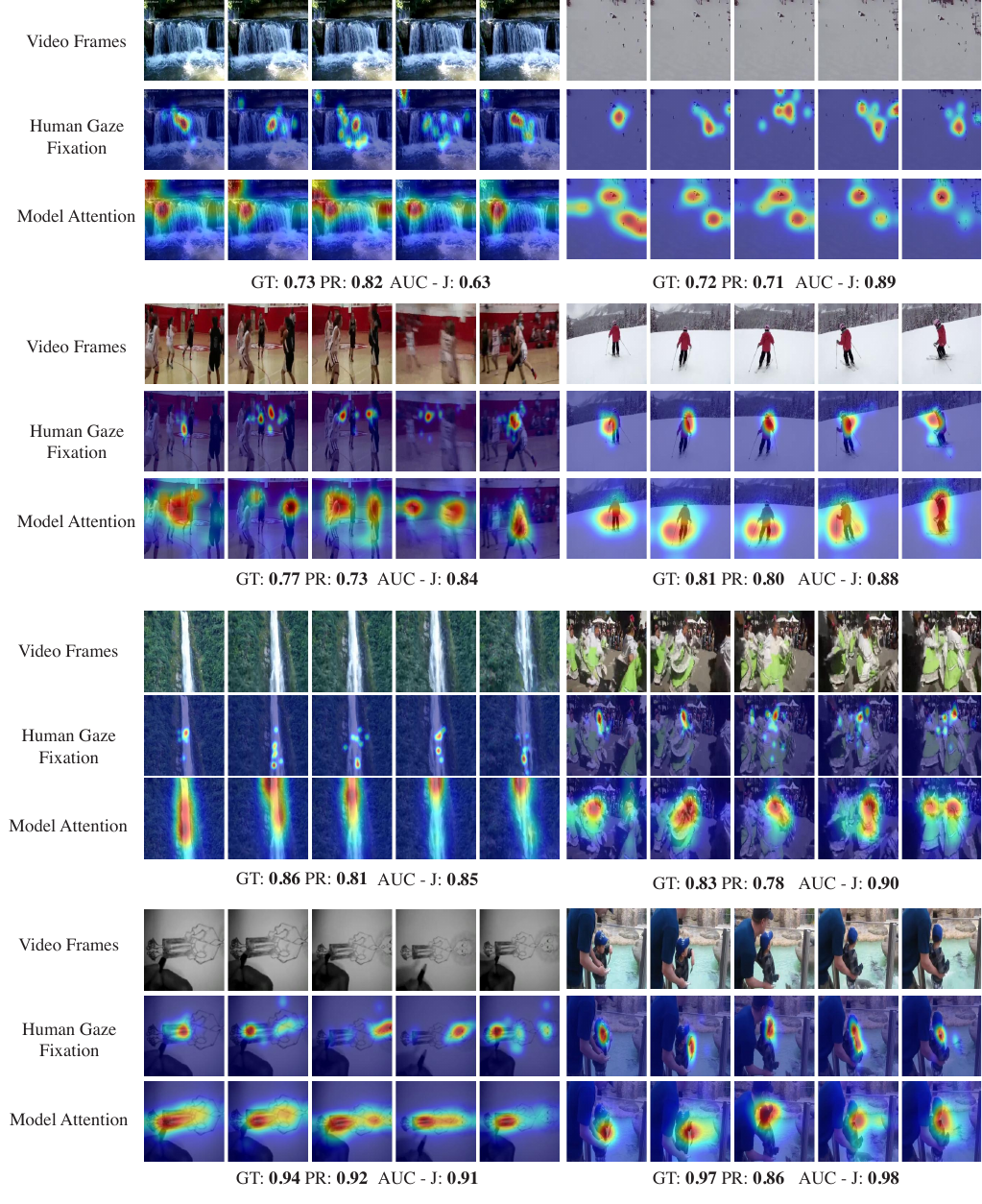}
\caption{Comparison of original video frames, gaze fixation maps, and model attention maps on the Memento10K dataset. We also indicate the ground-truth and predicted memorability scores, and the AUC Judd score measuring similarity between saliency maps.}
\label{fig:heatmaps_memento}
\end{figure*}

\begin{figure*}[t]
\centering
\includegraphics[width=\textwidth]{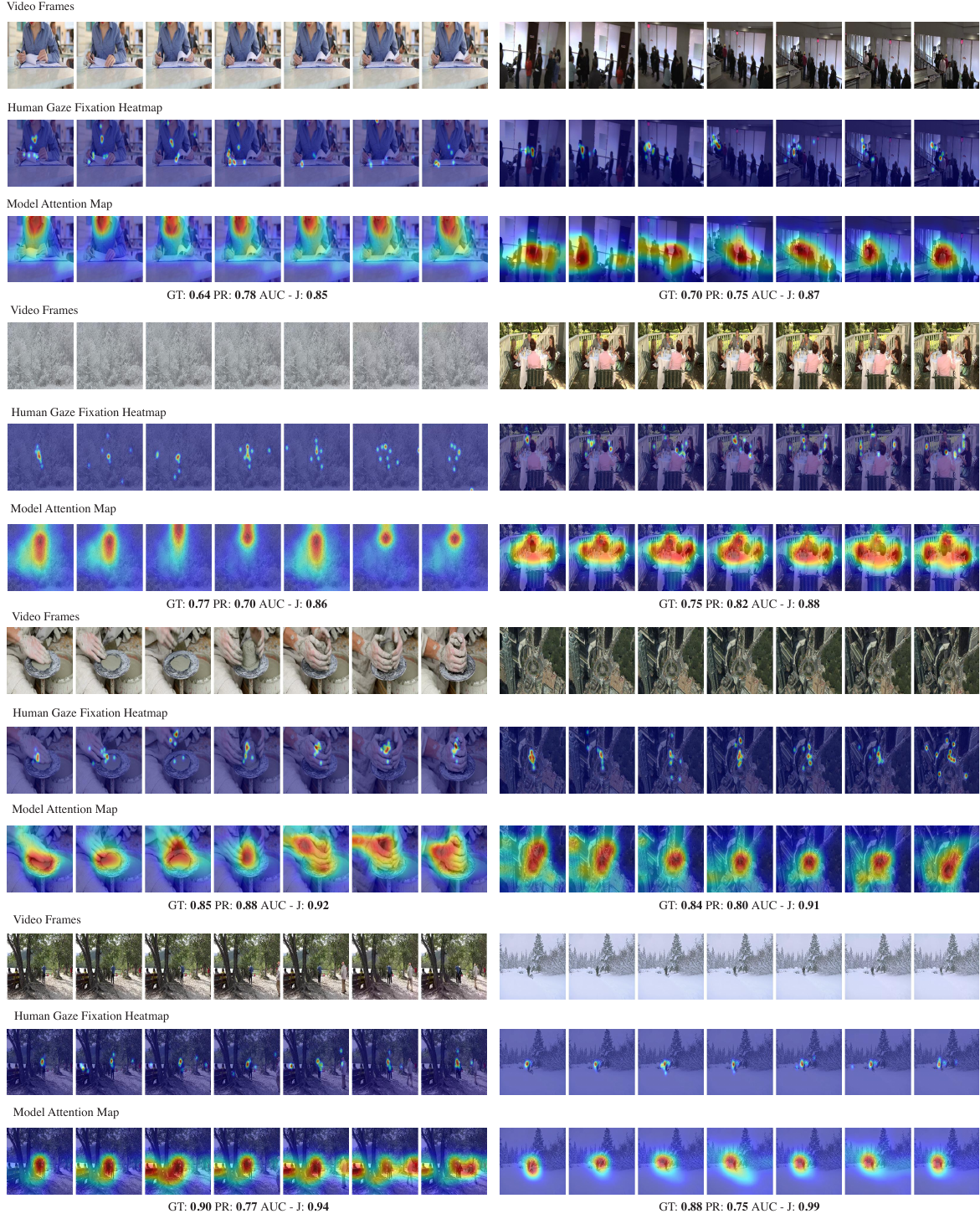}
\caption{Comparison of original video frames, gaze fixation maps, and model attention maps on the VideoMem dataset. 
We also indicate the ground-truth and predicted memorability scores, and the AUC Judd score measuring similarity between saliency maps.}
\label{fig:heatmaps_videomem}
\end{figure*}

\paragraph{Visualizing gaze and attention maps.}
The human gaze fixation maps and model attention maps across multiple videos can be seen in \cref{fig:heatmaps_memento} for Memento10k and \cref{fig:heatmaps_videomem} for VideoMem.
In both cases, model attention maps appear to be more similar to human gaze maps in higher memorability (GT) videos compared to lower memorability ones.
Note, in \cref{sec:attVGazeText}, we rule out the possibility that this alignment between model attention and human gaze is driven by center-bias.

\begin{figure*}[t]
\centering
\includegraphics[width=\textwidth]{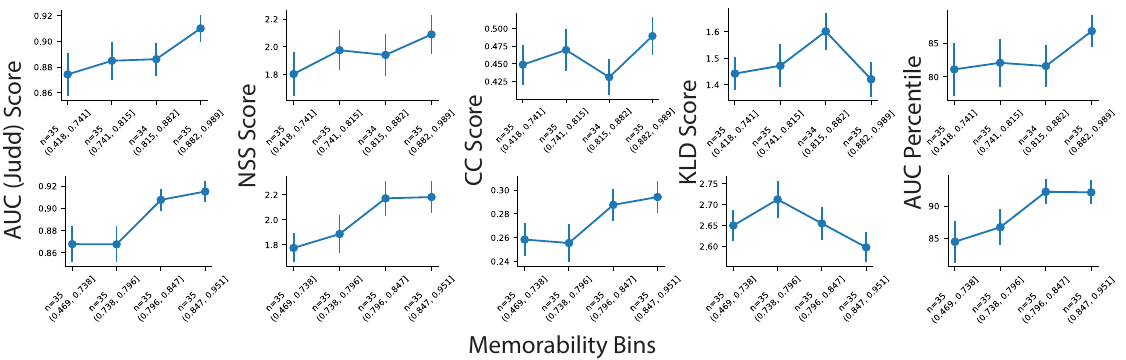}
% \vspace{-6mm}
\caption{\textit{Top:} Memento10K, \emph{Bottom:} VideoMem.
Performance across different similarity/distance metrics while comparing the human gaze fixation maps with model attention maps.
The metrics are indicated with arrows to indicate whether higher or lower scores are better:
AUC (Judd) $\uparrow$;
NSS $\uparrow$;
CC $\uparrow$;
KLD $\downarrow$; and
AUC Percentile (ours) $\uparrow$.
Results are presented for $n{=}139$ videos, binned into 4 percentiles based on ground-truth memorability scores.}
\label{fig:metrics}
% \vspace{-4mm}
\end{figure*}

\begin{table*}[t]
\centering
\small
\tabcolsep=0.10cm

\begin{tabular}{l c c c c c c c c c c c c}
\toprule
& \multicolumn{6}{c}{Memento10K} & \multicolumn{6}{c}{VideoMem} \\ 
\cmidrule(lr){2-7} \cmidrule(lr){8-13} 
Metrics & \multicolumn{3}{c}{M-H} & \multicolumn{3}{c}{H-H} & \multicolumn{3}{c}{M-H} & \multicolumn{3}{c}{H-H} \\
\cmidrule(lr){2-4} \cmidrule(lr){5-7} \cmidrule(lr){8-10} \cmidrule(lr){11-13}
& Simple & Complex & $t$ & Simple & Complex & $t$ & Simple & Complex & $t$ & Simple & Complex & $t$\\ 
\midrule
{\scriptsize AUC-J} $\uparrow$ & \meansem{0.89}{0.01} & \meansem{0.88}{0.01} & {\scriptsize$0.60$} & \meansem{0.90}{0.01} & \meansem{0.89}{0.01} & {\scriptsize$0.17$} & \meansem{0.89}{0.01} & \meansem{0.89}{0.01} & {\scriptsize$-0.19$} & \meansem{0.83}{0.01} & \meansem{0.80}{0.01} & {\scriptsize$1.53$ }\\
{\scriptsize AUC-P} $\uparrow$ & \meansem{84.41}{2.18} & \meansem{81.10}{2.48} & {\scriptsize$0.99$ }& - &  - & - & \meansem{89.45}{1.92} & \meansem{88.37}{1.72} & {\scriptsize$0.41$ }& - & - & -\\
{\scriptsize NSS} $\uparrow$ & \meansem{1.89}{0.09} & \meansem{1.60}{0.07} 
 & {\scriptsize$\textbf{2.1}$} & \meansem{3.02}{0.17} & \meansem{3.07}{0.17} & {\scriptsize$-0.20$} & \meansem{2.08}{0.14} & \meansem{1.94}{0.11} & {\scriptsize$1.05$ }& \meansem{3.85}{0.46} & \meansem{3.79}{0.40} & {\scriptsize$0.11$} \\
{\scriptsize CC} $\uparrow$ & \meansem{0.58}{0.02} & \meansem{0.52}{0.02} & {\scriptsize$1.66$ }& \meansem{0.48}{0.02} & \meansem{0.49}{0.02} & {\scriptsize$-0.26$} & \meansem{0.29}{0.02} & \meansem{0.26}{0.01} & {\scriptsize$1.46$ }& \meansem{0.28}{0.02} & \meansem{0.28}{0.01} & {\scriptsize$0.14$}\\
{\scriptsize KLD} $\downarrow$ & \meansem{1.08}{0.02} & \meansem{1.16}{0.02} & {\scriptsize$-1.41$} & \meansem{2.19}{0.11} & \meansem{2.16}{0.11} & {\scriptsize$0.23$ }&  \meansem{2.64}{0.05} & \meansem{2.67}{0.04} & {\scriptsize$-0.64$} & \meansem{4.01}{0.13} & \meansem{4.16}{0.10} & {\scriptsize$-0.89$}\\
\bottomrule
\end{tabular}
\vspace{2mm}
\caption{Comparing gaze fixation maps against model's attention map via different metrics for simple and complex videos, along with human-human alignment scores(split by half, averaged over 10 random iterations) for Memento10K and Videomem datasets. $\uparrow$ ($\downarrow$) indicates higher (lower) is better.
M-H: Model-human;
H-H: Human-human; and
$t$: t-test significance. Significant t-statistics are shown in bold ($p<0.05$).
}
\label{tab:simple_vs_complex_combined}
\vspace{-2mm}
\end{table*}

\subsection{Additional Results Comparing Human Gaze \vs~Model Attention (Video Memorability)}
\label{subsec:metrics_and_complexity}

We expand on the evaluation of human gaze and model attention alignment using additional metrics and explore how video complexity affects this alignment.

\paragraph{Additional similarity metrics.}
To compare human gaze fixation maps to the model's attention maps, we use standard metrics used in saliency evaluation such as AUC-Judd, NSS, CC, KLD. Additionally, we develop and apply a novel shuffle-based metric, the AUC-Percentile.

While \cref{fig:true_bins} from the main paper shows results only on AUC-Judd and NSS due to space restrictions, we now extend this to all metrics in \cref{fig:metrics}.
We observe a common trend of greater match between human gaze and model attention maps with increasing memorability scores across most metrics, indicating that memorable videos attract both human and model attention to the same regions of the video frames.

\paragraph{Impact of video complexity on gaze/attention alignment} We split the videos in each dataset at the median of the average number of objects per frame to get one group of simpler and one group of more complex videos. We computed model attention-human gaze (M-H) and human-human (H-H) gaze alignment scores for these groups of videos.  The alignment metrics are presented in \cref{tab:simple_vs_complex_combined} and indicate that in both datasets, humans gaze patterns tend to agree with those of other humans as well as model attention patterns with no statistically significant differences between simple and complex videos except in the M-H NSS metric for Memento10k. Therefore, the results presented in the main paper are unlikely to be explained by complexity of the videos.

% \PK{Furthermore, to investigate the impact of cognitive load on the alignment of human fixations with model attention maps, we analyzed simple and complex videos in our dataset. These categories were defined by splitting the videos in each dataset at the median of the average number of objects per frame. The similarity metrics for this analysis are presented in \cref{tab:simple_vs_complex_combined}. Our results indicate that in the Me-
% mento Dataset, humans tend to simple and complex videos
% similarly, while there is a slightly better alignment in simple
% videos for the VideoMem dataset. The overall performance
% of our model shows slightly better alignment for simpler
% videos across both datasets.}

\begin{figure*}[t]
\centering
\includegraphics[width=\textwidth]{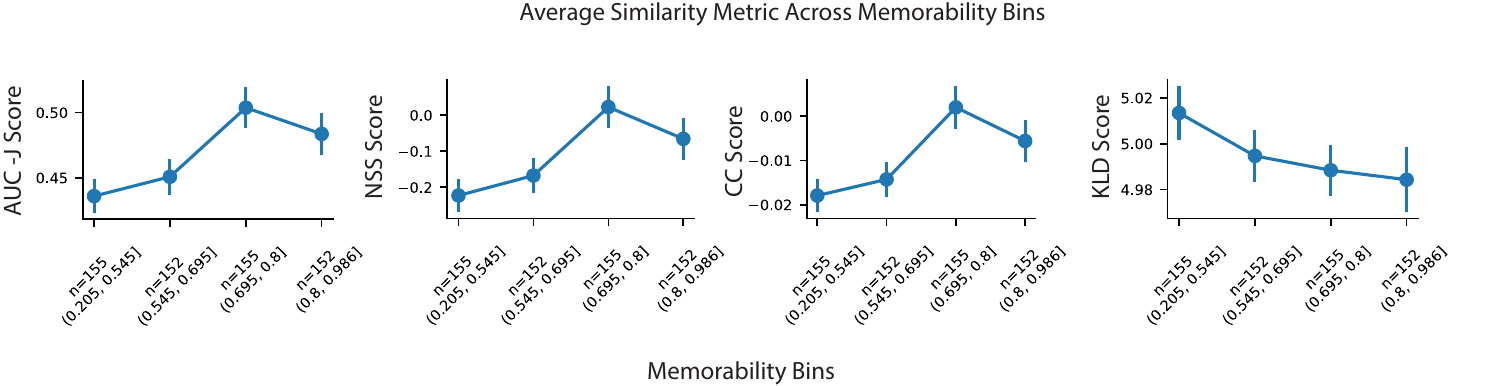}
% \vspace{-6mm}
\caption{
Performance across different similarity/distance metrics while comparing the human gaze fixation maps with model attention maps for the FIGRIM dataset.
The metrics are indicated with arrows to indicate whether higher or lower scores are better:
AUC (Judd) $\uparrow$;
NSS $\uparrow$;
CC $\uparrow$; and
KLD $\downarrow$.
Results are presented for $n{=}614$ images, binned into 4 percentiles based on ground-truth memorability scores.}
\label{fig:FIGRIM_metrics}
% \vspace{-4mm}
\end{figure*}

\subsection{Human Gaze \vs~Model Attention (Image Memorability)}
\label{subsec:image_memorability}
Next, to establish the general trend of similarity between model attention and human gaze with increasing memorability, we also present results on the FIGRIM dataset, which provides gaze data along with memorability scores for images.
While \cref{sec:imagetovideo} provides quantitative results on memorability prediction,
\cref{fig:FIGRIM_metrics} illustrates a similar trend of increasing human gaze and model attention agreement with increasing memorability scores on the FIGRIM dataset.

\section{Modeling with Captions}
\label{sec:methodExt}
Building upon \cref{subsec:cnn_transformer}
where we presented the vision-only model, we now explain how captions can be easily integrated into the existing  modeling framework.
We consider two paradigms.
In the first, the caption is assumed available, both during training and inference.
This may be achieved using recent advances in vision-language models (VLMs).
In the second, we consider experiments where the caption is predicted simultaneously with the estimation of the video memorability score (similar to~\cite{Newman2020}).

\subsection{Assuming Caption is Available}

When the caption is given, we first extract token-level representations through a BERT encoder and append them to the spatio-temporal video tokens for memorability prediction.

\paragraph{Text encoder.}
We extract textual embeddings for the captions from the last hidden state of the BERT~\cite{devlin2019bert} model $\psi$:
\begin{equation}
\{ \bg_l \}_{l=1}^N = \psi(\{ g_l \}_{l=1}^N) \, , 
\end{equation}
where $\bg_l \in \bbR^{d}$, $N$ is the number of tokens, and $d$ is the dimensionality of the embeddings, equal to the reduced dimensionality of images after the linear layer.

\paragraph{Changes to the video encoder.}
We append $N$ text tokens to the $THW$ visual tokens fed to the Transformer encoder.
To distinguish between text and image, we append modality specific embeddings to both the visual (from
Eq.~2)
% \cref{eq:pos_embed})
and text tokens. 
We also add position embeddings indicating order to the text tokens.
\begin{align}
\bbf'_{ij} &= \bW^d \bbf_{ij} + \bE^t_i + \bE^s_j + \bE^m_1 \, , \\
\bg'_l &= \bg_l  + \bE^c_l + \bE^m_2 \, ,
\end{align}
where $i = [1, \ldots, T]$,
$j = [1, \ldots, HW]$,
$l = [1, \ldots, N]$,
$\bE^t_i$ is the $i^\text{th}$ row of the temporal embedding matrix (learnable or Fourier) for images, 
$\bE^c_l$ is the $l^\text{th}$ row of the temporal embedding matrix for the caption,
$\bE^s_j$ is the $j^\text{th}$ row of the spatial embedding matrix, and
$\bE^m_{[1,2]}$ are the modality embeddings, one for visual tokens, another for text.

We combine the $\CLS$ token (with learnable parameters $\bh_\CLS$), image and text tokens to create a sequence of $1+TWH+N$, apply LayerNorm, feed it to the TE. 
\begin{multline}
[\tilde{\bh}_\CLS, \tilde{\bbf}_{11}, \ldots, \tilde{\bbf}_{THW}, \tilde{\mathbf{g}}_1, \ldots, \tilde{\mathbf{g}}_N] = \\
\text{TE}( [\bh_\CLS, \bbf'_{11}, \ldots, \bbf'_{THW}, \mathbf{g}'_1, \ldots, \mathbf{g}'_N ]) \, .
\end{multline}
As before, $\tilde{\bh}_\CLS$ is used to predict the memorability score.

We report results when using the ground-truth caption in this approach in
% \cref{tab:text_ablation},
Tab.~1,
row 7 of the main paper (w original captions as input).
For Memento10k, we see a 0.04 points increase in Spearman correlation (0.706 to 0.745), however, captions do not seem to assist VideoMem.

\subsection{Joint Prediction of Caption and Memorability}

When the caption is not available, we consider predicting the caption along with the memorability scores.
In particular, we adapt CLIPCap~\cite{mokady2021clipcap}, a recent approach that connects CLIP visual features with the GPT-2 decoder using a Transformer mapping layer.

Specifically, we use a mapping network (a Transformer decoder) to convert the $THW$ visual tokens at the output of the Transformer encoder $\tilde{\bbf}_{ij}$ to a set of $P$ prefix tokens.
The mapping network of $L_D{=}6$ layers consists of $P$ query learnable tokens and uses visual inputs as memory, $P{=}30$.
The outputs of this mapping network are fed as prefix tokens to the GPT-2, and captions are generated in an auto-regressive manner.

We train the model jointly, to predict both the memorability score (using L1 regression loss) and the caption (using cross-entropy loss).
Results of this approach are presented in 
% \cref{tab:text_ablation},
Tab.~2,
row 3.
A small increase of 0.004 is observed in the RC score (0.706 to 0.710) for Memento10k, while VideoMem continues to not benefit from captions.

We conclude that generating captions separately with a VLM and using them (as shown above) may be a better course of action than training a joint model.

\section{Panoptic Segmentation}
\label{sec:panoptic_supp}

We present additional experiments and results from the semantic \emph{stuff} \vs~\emph{things} analysis obtained through panoptic segmentation. 

\begin{figure*}[t]
\centering
\includegraphics[width=0.9\textwidth]{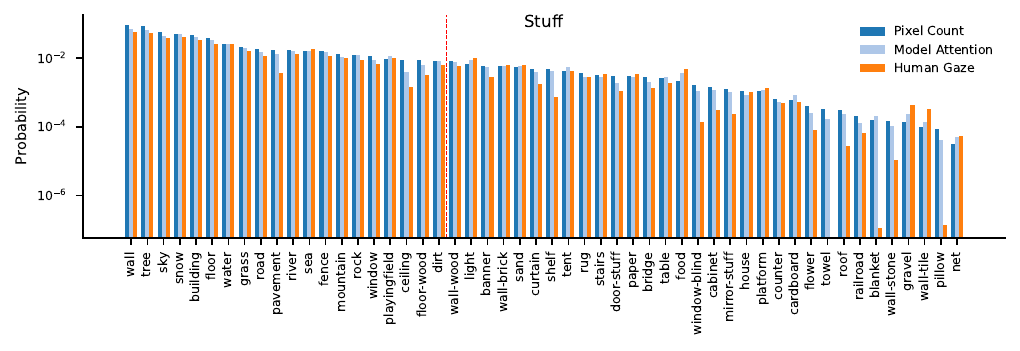}
\includegraphics[width=0.9\textwidth]{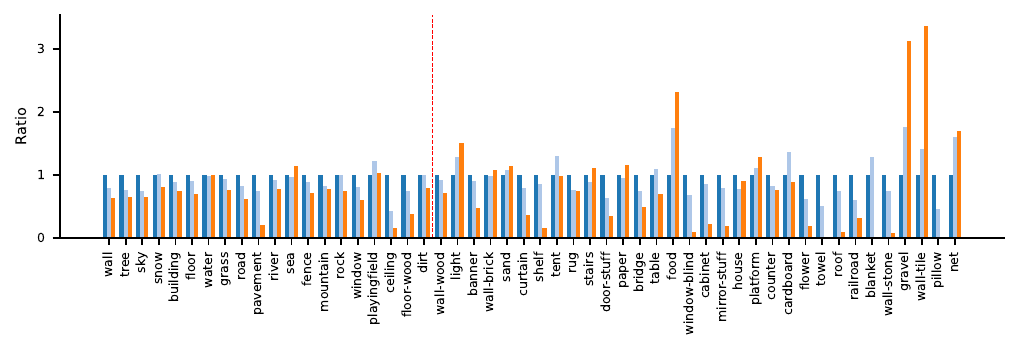}
\includegraphics[width=0.9\textwidth]{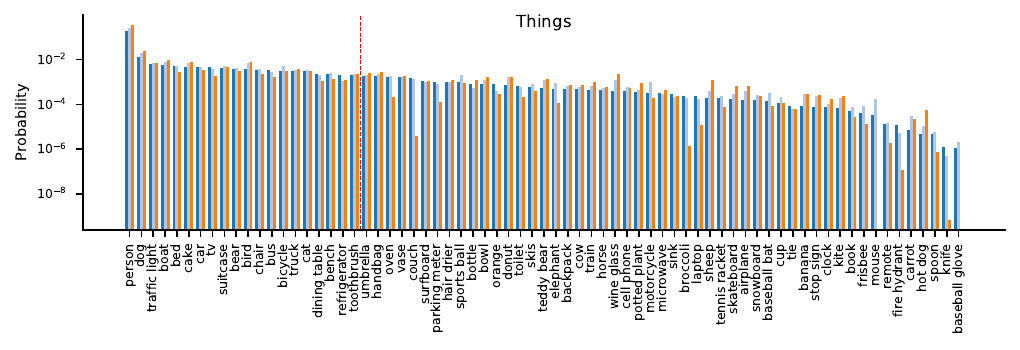}
\includegraphics[width=0.9\textwidth]{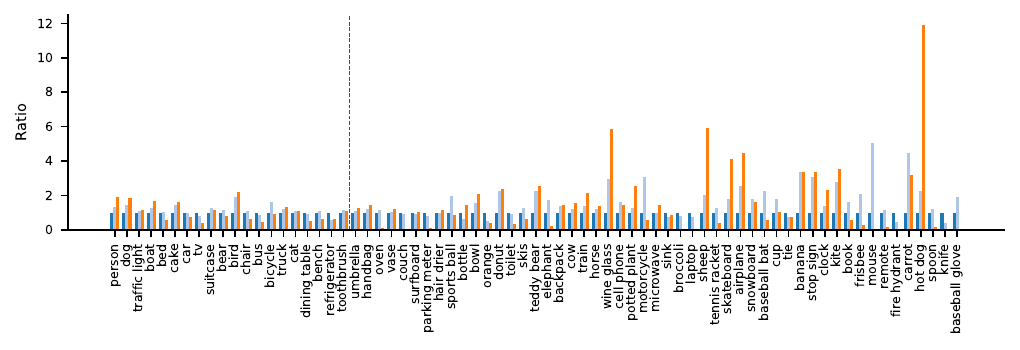}
\caption{Analysis of panoptic segmentation results.
The vertical red line marks the top-20 labels within these categories.
\textbf{First} and \textbf{Third}:  Raw, attention-, and gaze-weighted pixel probabilities for \textit{stuff} and \textit{things}, respectively (plotted in semilog scale);
\textbf{Second} and \textbf{Fourth}: Highlights how model attention-weighted and human gaze-weighted pixel counts are higher or lower relative to normalized raw pixel counts for \textit{stuff} and \textit{things}.}
\label{fig:supp_panoptic_dist}
\end{figure*}

\paragraph{Pixel count, human gaze, and model attention across all labels.}
In \cref{fig:supp_panoptic_dist}, we show the distributions for all stuff and things labels. 
Row 1 is the probability distribution of pixel counts and gaze/attention weighted counts for stuff labels (plotted in semilog scale). 
In row 2, we normalize these counts by the pixel count (blue), highlighting dynamic stuff labels such as \emph{light, food, platform} receiving higher attention weighted scores, while other mundane labels such as \emph{wall, sky, road} receiving lower scores.

A similar analysis is shown for \emph{things} in rows 3 and 4.
Here too, we observe that daily objects such as \emph{bed, car, toilet} receive less human and model attention to account for memorability, while dynamic or interesting objects such as \emph{person, dog, bird, wine glass, banana} (among others) receive higher attention.
This confirms that not all objects are interesting.

\begin{figure*}[t]
\centering
\includegraphics[width=0.9\textwidth]{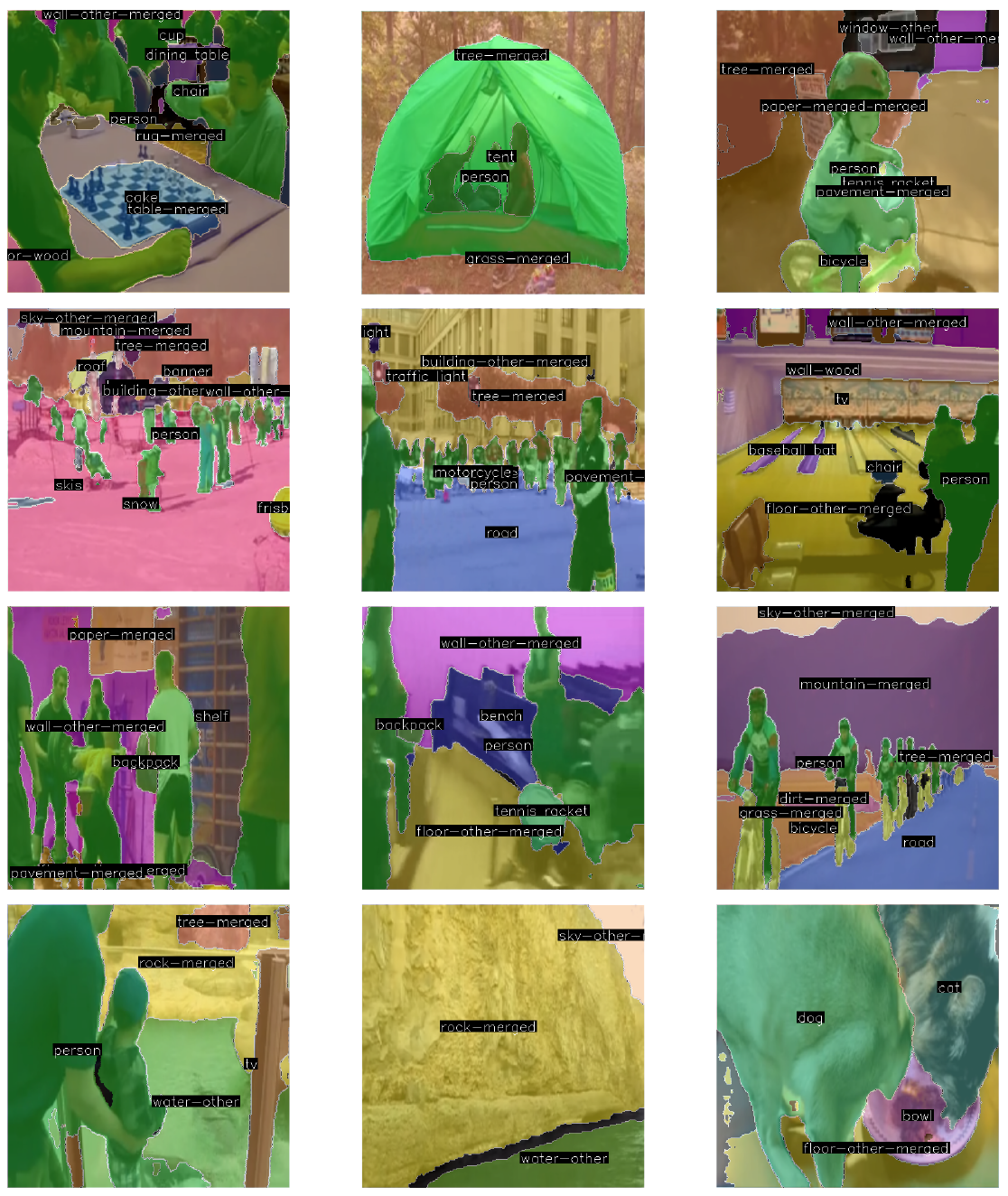}
\caption{Visualisation of panoptic segmentation predictions on Memento10k dataset. }
\label{fig:panoptic_all}
\end{figure*}

Note, while this analysis is also subject to accuracy of Maskformer~\cite{maskformer} (the panoptic segmentation approach),
qualitatively, we find this to be quite reliable as seen in \cref{fig:panoptic_all}.

% The later labels (lower occurrence probability) exhibit large deviations in attention. Specifically, none of the leading 20 \emph{stuff} labels show an attention increase exceeding 1.5 times (Group +).
% To capture a broader spectrum of attentional variations, we expanded our analysis to encompass all labels. 

% \paragraph{Memorability \vs~Attention density.}
% Recall, we test whether specific labels are related to memorability, by splitting all labels into three attention pattern groups:
% \textbf{G+}: attention/gaze increased by over 1.5$\times$ compared to the pixel counts;
% \textbf{G=}: attention/gaze did not change appreciably (between 0.8$\times$ and 1.2$\times$ the pixel counts); and 
% \textbf{G-}: attention/gaze decreased significantly (falling below
% 0.5$\times$ the pixel counts).
% While
% % \cref{fig:stuff_density}
% Fig.~6
% of the main paper and related discussion shows significant relationship between some stuff labels and memorability, we do not see significant results ($p < 0.05$) for \emph{things}, as seen in \cref{fig:things_density}.

\end{document}